\documentclass[journal]{IEEEtran}

\usepackage{graphicx}
\usepackage{multirow}
\usepackage{titlesec}
\usepackage{booktabs}
\usepackage[table]{xcolor}
\usepackage{soul}
\usepackage{url}

\ifCLASSINFOpdf
\else
\fi
\usepackage[inline]{enumitem}
\usepackage[linesnumbered,ruled,vlined]{algorithm2e}
\usepackage{amsmath}
\usepackage{subfig}
\usepackage{graphicx}
\usepackage{float}
\usepackage{siunitx}
\usepackage{xcolor}
\usepackage{gensymb}
\usepackage{cite}
\usepackage{amsmath,amssymb,amsfonts}
\usepackage{algorithmic}
\usepackage{textcomp}
\usepackage[inline]{enumitem}
\usepackage[linesnumbered,ruled,vlined]{algorithm2e}
\usepackage{float}
\usepackage{booktabs}
\usepackage{array, booktabs, makecell}

\SetKwInput{KwInput}{Input}                
\SetKwInput{KwOutput}{Output}              
\SetKwRepeat{Repeat}{do}{until}

\begin{document}
%
\title{Federated Hyperdimensional Computing}
%
%
%

\author{Kazim Ergun*, Rishikanth Chandrasekaran*,       and~Tajana~Rosing\\
\textit{Department of Computer Science and Engineering, University of California San Diego}

\thanks{\textsuperscript{*}Both authors contributed equally to this research}

}

\maketitle



\begin{abstract}
Federated learning (FL) enables a loose set of participating clients to collaboratively learn a global model via coordination by a central server and with no need for data sharing. Existing FL approaches that rely on complex algorithms with massive models, such as deep neural networks (DNNs), suffer from computation and communication bottlenecks. 
In this paper, we first propose FedHDC, a federated learning framework based on hyperdimensional computing (HDC). FedHDC allows for fast and light-weight local training on clients, provides robust learning, and has smaller model communication overhead compared to learning with DNNs. 
However, current HDC algorithms get poor accuracy when classifying larger \& more complex images, such as CIFAR10.  To address this issue, we design FHDnn, which complements FedHDC with a self-supervised contrastive learning feature extractor.  We avoid the transmission of the DNN and instead train only the HDC learner in a federated manner, which accelerates learning, reduces transmission cost, and utilizes the robustness of HDC to tackle network errors. We present a formal analysis of the algorithm and derive its convergence rate both theoretically, and show experimentally that FHDnn converges 3$\times$ faster vs. DNNs. The strategies we propose to improve the communication efficiency enable our design to reduce communication costs by 66$\times$ vs. DNNs, local client compute and energy consumption by ~1.5 - 6$\times$, while being highly robust to network errors.  Finally, our proposed strategies for improving the communication efficiency have up to 32$\times$ lower communication costs with good accuracy.


\end{abstract}



\vspace{-5mm}

%
\IEEEpeerreviewmaketitle

\section{Introduction}

Recent years have witnessed an unprecedented growth of data sensing and collection by the Internet of Things (IoT). It is estimated that the number of interconnected IoT devices will reach 40 billion by 2025, generating more than 79 zettabytes (ZB) of data~\cite{IDC}. Empowered by this massive data, emerging Deep Learning (DL) methods enable many applications in a broad range of areas including computer vision, natural languag processing, and speech processing~\cite{lecun2015deep}.

In the traditional cloud-centric DL approach, data collected by
remote clients, e.g. smartphones, is gathered centrally at a computationally powerful server or data center, where the learning model is trained. Often, the clients may not be willing to share data with the server due to privacy concerns. Moreover, communicating massive datasets can result in a substantial burden on the limited network resources between the clients and the server. This motivated the development of distributed algorithms to allow machine learning at edge networks without data sharing. Federated learning (FL), proposed in~\cite{mcmahan2017communication}, has recently drawn significant attention as an alternative to centralized learning. FL exploits the
increased computational capabilities of modern edge devices to
train a model on the clients’ side while keeping
their collected data local. In FL, each client performs model training based on its local dataset and shares the model with a central server. The models from all participating clients are then aggregated to a global model.

Learning in FL is a long-term process consisting of many progressive rounds of alternating computation and communication. Therefore, two of the main challenges associated with FL are the computation and communication bottlenecks~\cite{kairouz2021advances}. With FL, the computation, i.e. the training process, is pushed to edge devices. However, state-of-the-art ML algorithms, including deep
neural networks (DNN), require a large amount of computing power and memory resources to provide better service quality. The DNN models have complicated
model architectures with millions of parameters and require backpropagation, resulting in prohibitively long training times. Besides computation, the communication load of DNN based FL suffers from the need to repeatedly convey massive model parameters between the server and large number of clients over wireless networks~\cite{li2020federated}.

Another challenge arises when FL is carried out over wireless networks. The wireless channels are unreliable in nature, introducing noise, fading, and interference to the transmitted signals. Therefore, the communication in wireless FL is prone to transmission errors. The common solution for this problem is using multiple-access technologies~\cite{goldsmith2005wireless} (e.g., TDMA, OFDMA) to prevent interference and error-correcting codes to overcome noise. If there still exists any errors, then a reliable transport layer protocol~\cite{kurose2005computer} (e.g.. TCP) is adopted, where acknowledgment, retransmission, and time-out mechanisms are employed to detect and recover from transmission failures. This reliability comes with a price; achieving error-free communication requires a lot of wireless resources, increases energy consumption, limits communication rates, and hence decreases the training speed and convergence of FL. Otherwise, in an unreliable scenario, the transmission errors will impact the quality and correctness of the FL updates, which, in turn, will affect the accuracy of FL, as well as its convergence.


This paper proposes a novel technique that enables
efficient, robust, and accurate federated learning using brain-inspired models in high-dimensional space. Instead of conventional machine learning algorithms, we exploit Hyperdimensional Computing (HDC) to perform lightweight learning with simple operations on distributed low-precision vectors, called hypervectors. HDC defines a set of operations to manipulate these hypervectors in the high-dimensional vector space, enabling a computationally tractable and mathematically rigorous framework for learning tasks. A growing number of works have applied HDC to a wide range of learning problems, including reasoning~\mbox{\cite{kanerva2010we}}, biosignal processing~\mbox{\cite{asgarinejad2020detection}}, activity prediction~\mbox{\cite{imani2019semihd}}, speech/object recognition~\mbox{\cite{imani2018hierarchical,imani2017voicehd}}, and prediction from multimodal sensor fusion~\mbox{\cite{rasanen2015sequence}}. These studies have demonstrated the high efficiency, robustness and effectiveness of HDC in solving various learning problems, highlighting its potential as a powerful tool for a variety of applications.


HDC has various appealing characteristics, particularly for edge devices. It is well-suited to address the challenges in FL as: \begin{enumerate*}[label=(\roman*)]
    \item HDC is low-power, computationally efficient, and amenable to hardware-level optimization~\cite{khaleghi2022generic},
    \item it is fault tolerant, providing strong robustness in the presence of errors~\cite{morris2022HyDREA},
    \item HDC models are small, thus both memory-efficient and communication-efficient~\cite{khaleghi2021tiny}, 
    \item HDC encoding can transform non-linear learning tasks into linear optimization problems~\cite{thomas2021theoretical}, and
    \item HDC enables fast and light-weight learning with its simple operations~\cite{khaleghi2021tiny}.
\end{enumerate*} These features make HDC a promising
solution for FL using today’s IoT edge devices with constrained storage,
battery, and resources, over wireless networks with latency concerns and limited bandwidth.

We address several technical challenges to enable \textit{federated hyperdimensional computing} at the IoT edge. Although HDC is inherently suitable for FL, current HDC algorithms fail to provide acceptable accuracy for complex image analysis~\cite{imani2019bric,zou2021manihd}, which is one of the key FL applications. Recently published work~\cite{dutta2022hdnn} combines convolutional neural networks (CNNs) with HDC to learn effectively on complex data. It leverages convolution-based feature extraction prior
to the HD encoding step. Unfortunately, such a configuration (DNN+HDC) possesses the aforementioned computation and communication drawbacks of DNNs for FL. The other challenge  is the communication of HDC models over unreliable wireless channels. While the robustness of HDC encoded data to noise and bit errors was demonstrated by prior work~\cite{morris2022HyDREA,thomas2021theoretical}, similar claims were not investigated for an entire HDC model itself. Finally, HDC models have a lot of redundancy that still can put a burden on communication efficiency, even though they are much smaller than DNN models.

SecureHD \mbox{\cite{8814566}}, HDnn \mbox{\cite{10.1145/3526241.3530331}} and our work, FedHDC, are all approaches that leverage high-dimensional computing for various tasks. However, there are distinct differences between these methods. FedHDC focuses on federated learning, which enables collaborative model training across multiple decentralized devices while maintaining data privacy. In contrast, SecureHD \mbox{\cite{8814566}}, emphasizes secure high-dimensional computing, specifically designed to handle classification tasks with a focus on security.

HDnn \mbox{\cite{10.1145/3526241.3530331}}, on the other hand, is a hybrid approach that combines high-dimensional computing with convolutional neural networks (CNNs). This method aims to harness the strengths of both HDC and CNNs to improve classification performance, particularly in image recognition tasks. While FHDnn and HDnn both utilize high-dimensional computing, the primary difference lies in their objectives: FedHDC targets federated learning, whereas HDnn \mbox{\cite{10.1145/3526241.3530331}} focuses on enhancing classification performance by integrating HDC with deep learning techniques. Unlike our proposed FHDnn, HDnn \mbox{\cite{10.1145/3526241.3530331}} trains the feature extractor to learn representations amenable to learning in the high-dimensional vector space.

In this paper, we first present federated hyperdimensional computing, FedHDC, an extension of the HDC paradigm into the federated learning domain.  Next we design a novel synergetic FL framework, called FHDnn, that enables FedHDC to also perform complex image classification by combining contrastive learning framework as a feature extractor with FedHDC, but still keeping model updates to only HDC portion, resulting in fast and accurate model updates via federated learning.  In the following, we summarize the main contributions of the paper:

\textit{i)} We present FedHDC to address the computation and communication problems in standard DNN based FL approaches. The simple and highly efficient operations of HDC allow for fast and low-weight local training on clients between communication rounds. FedHDC incurs very low communication overhead as HD models are very small in size and training requires many fewer rounds of communication to converge compared to DNNs, resulting in at least 66$times$ lower communication overhead as we shown in our results.

\textit{ii)} We analyze HDC training process using the language of gradient methods from statistical learning and optimization. This viewpoint helps us provide a formal treatment of FedHDC as a general framework for federated learning, and precisely study its convergence properties. FedHDC can achieve $\mathcal{O}(\frac{1}{T})$ convergence rate, with T representing the number of communication rounds, but such claim is not possible for non-convex and non-linear DNNs. As HD encoding embeds data into a high-dimensional space and can transform non-linear distributed learning tasks into linear optimization, FedHDC enjoys simpler training and faster convergence compared to DNNs as it uses only HD computing, while having the superior performance properties of non-linear models.




\begin{figure}
    \centering
    \includegraphics[width = 8.5cm]{"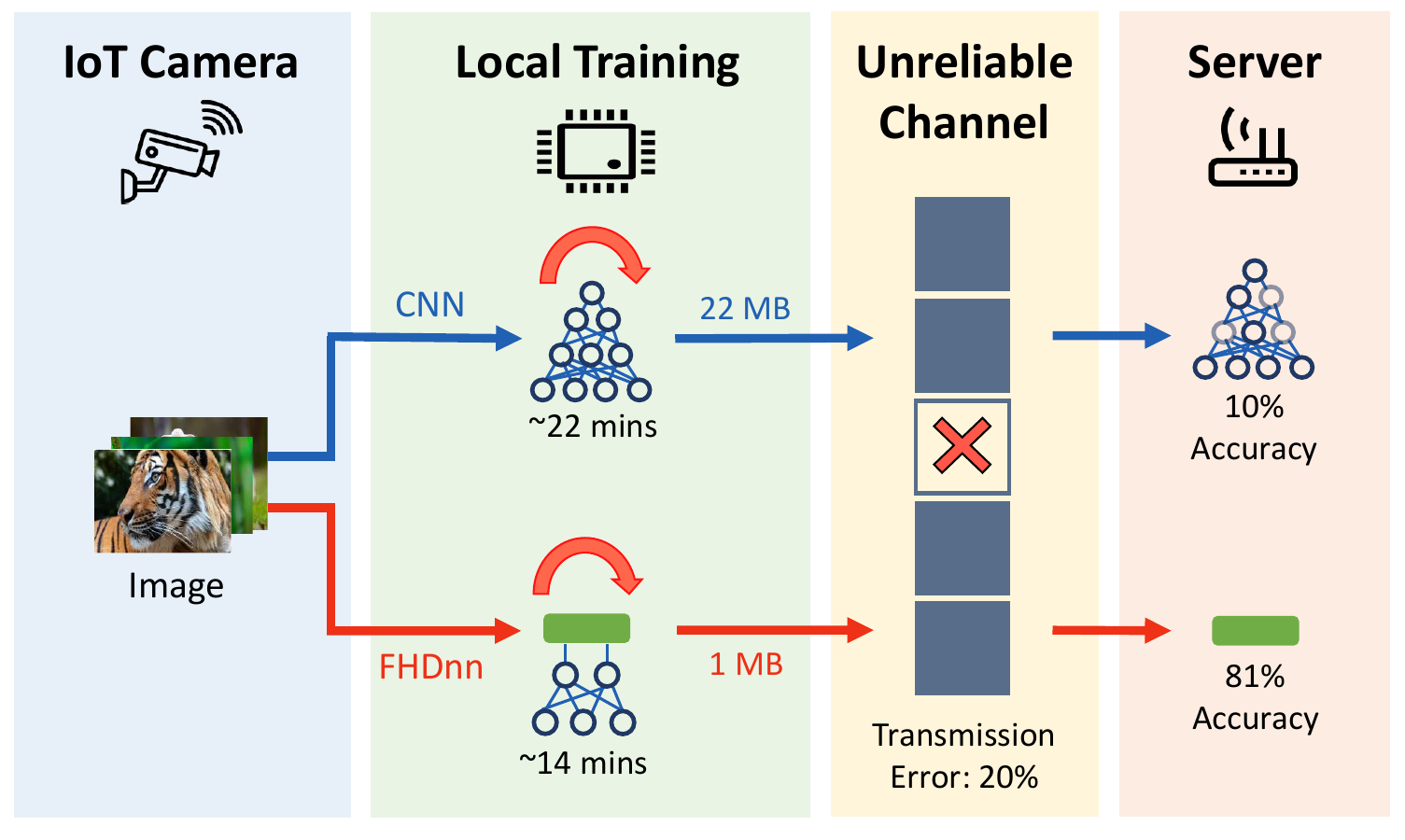"}
    \caption{FHDnn against CNNs for federated learning}
    \vspace{-1.5em}
    \label{fig:teaser}
\end{figure}

\textit{iii)} We present FHDnn, a novel synergetic FL framework that combines pre-trained CNN as a feature extractor with HDC. Specifically, we utilize a CNN trained using SimCLR \cite{chen2020simple}, a contrastive learning framework which learns informative representations of data self-supervised. FHDnn avoids the transmission of the CNN and instead trains only the HD learner in a federated manner. This strategy accelerates learning, reduces transmission costs, and utilizes the robustness of HDC to tackle network errors as shown in Fig.~\ref{fig:teaser}.




\textit{iv)} HD-based federated learning provides reliability for learning over unreliable wireless network with no additional cost. Unlike existing FL approaches, there is no need for multiple-access technologies to prevent interference or error-protection on the transmitted models. Due to such techniques, FL can have very limited communication rates, and hence low training speeds. We leverage the robustness of HDC and allow errors during transmission instead of limiting the rate to achieve error-free communication. We analyze FHDnn under three different unreliable network settings: packet loss, noise injection, and bit errors, and show that the perturbations in the client models can be tolerated by the HDC learner. A quantizer method with scaling is additionally proposed to enhance the resilience to bit errors.

\textit{v)} We also propose various strategies to further improve the communication efficiency of FedHDC and FHDnn. The HDC models have redundancy which we exploit to reduce their sizes for more efficient communication. We examine three approaches: binarized differential transmission, subsampling, and sparsification \& compression. We show their trade-offs between performance and efficiency through experiments.

We evaluate HDC-based federated learning by numerical experiments on different benchmark datasets and compare their performance with CNN based FL under various settings. We both theoretically and empirically show that the proposed approaches are robust to lossy network conditions. Based on our evaluations, FHDnn converges 3x faster than CNN, reduces the communication costs by 66$\times$ and the local computation cost on the clients by up to 6$\times$.  The communication efficiency of FedHDC and FHDnn is further improved by various strategies up to 32$\times$ with minimal loss in accuracy.

\section{Related Work}

Communication and computation bottlenecks of FL have been widely studied in the literature and various solutions were proposed targeting improvement at different parts of the overall process. FL involves many rounds of communication with the participation of numerous clients, typically at low rates over wireless links. These considerations have led to a significant interest in communication-efficient design of FL systems. Previous research has primarily focused on decreasing the size of the model updates~\cite{konevcny2016federated2} and reducing the number of communication rounds or communicating clients~\cite{reisizadeh2020fedpaq}. In addition, during each round of communication, participating clients train models locally on device for multiple epochs. Deep learning models that are commonly used tend to be expensive to train requiring backpropogation algorithm which is compute heavy. Efficient computation is also of great importance as clients are usually not equipped with powerful hardware. This is addressed in prior work by reducing the model complexity to alleviate local training~\cite{jiang2022model}. On the other hand, there is often a trade-off between communication and computation; one strategy for lowering the frequency of communication is to put more emphasis on computation. The lightweight nature of HDC models make them suitable for running on edge devices with constrained resources.

\subsection{Communication Efficiency}

A prototypical FL approach named FedAvg~\cite{mcmahan2017communication} enables flexible communication and computation trade-off. The work follows from the seminal research in distributed stochastic gradient descent (SGD). Improvement in communication-efficiency is achieved by allowing for the clients to run multiple local SGD steps per communication round.  Many succeeding studies have pursued the theoretical understanding of FedAvg in terms of communication-computation trade-offs and have carried out rigorous analysis of the convergence behavior depending on the underlying assumptions (e.g., IID or non-IID local datasets, convex or non-convex loss functions, gradient descent or stochastic gradient descent)~\cite{li2019convergence,karimireddy2020scaffold}. Another approach that directly affects local training is to modify model complexity. Some examples are pruning~\cite{zhu2017prune}, restricting the model weights to be numbers at a certain bitwidth~\cite{courbariaux2015binaryconnect}, and bounding the model size\cite{oktay2019scalable}. These methods also lower computation complexity along with communication overhead.

As the models for FL can get very large---especially in the case of DNNs---a different line of work explored methods to reduce the communicated model (or gradient) size, without altering the original local models. Existing schemes typically perform a form of compression, that is, instead of transmitting the raw model/gradient data, one transmits a compressed representation with fewer bits, for instance by means of limiting bitwidth (quantization) or enforcing sparsity (sparsification). Particularly, a popular class of quantization operators is based on random dithering~\cite{horvath2019natural}. Sparsification methods decrease the number of non-zero entries in the communicated data to obtain sparse vectors~\cite{wangni2018gradient}. Structured and sketched updates are also proposed in~\cite{konevcny2016federated}, which can be further supported by lossy compression and federated dropout~\cite{caldas2018expanding}. Some other approaches include randomized techniques such as stochastic rounding~\cite{suresh2017distributed}, subsampling~\cite{konevcny2016federated2}, and randomized approximation~\cite{konevcny2018randomized}.

In FL, a group of clients might often provide similar, and hence redundant, model information during communication rounds. Orthogonal to the compression-based approaches, one can dismiss the updates of some clients as communicating all model updates would be an inefficient use of resources. Early works have attempted simple client selection heuristics such as selecting clients with higher losses ~\cite{balakrishnan2020resource}, sampling clients of larger update norm with higher probability~\cite{chen2020optimal}, and sampling clients with probabilities proportional to their local dataset size~\cite{mcmahan2017communication}, but  the similarity or redundancy of the client updates are not exploited in these methods. Ideally, a diverse and representative set of clients should be selected that contribute different, informative updates. In consideration of this, several selection criteria have been investigated in recent literature, some of which are diversity-based selection~\cite{balakrishnan2021diverse}, importance sampling~\cite{chen2020optimal}, and selection by update significance~\cite{chen2018lag}.

FL is often carried out over wireless channels that attenuate the transmitted signal as well as introduce noise, and thus the communication is unreliable, prone to transmission errors. All the aforementioned approaches assume reliable links and ignore the wireless nature of the communication medium. The inherent assumption is that independent error-free communication ``tunnels'' has been established between the clients and the server by some existing wireless protocol.  A common way to achieve this is to divide the channel resources among clients with multiple-access technologies (e.g., TDMA, CDMA, OFDMA) to mitigate interference, and utilize powerful error correcting codes to overcome noise~\cite{gupta2015survey}. However, the communication rates and consequently the overall training speed suffer due to the limited channel resources that can be allocated per client.


\subsection{Computation Efficiency}

The clients in FL are typically resource-constrained, battery-operated edge devices with limited power and computation budgets, unlike the powerful servers used in cloud-centric learning. DNN-based FL methods require clients to perform on-device backpropagation during each round of training which is computationally expensive and is incurring high resource usage. To overcome this challenge, prior works mainly explored low complexity NN architectures and lightweight algorithms suitable for edge devices. A lot of the `local methods' for improving communication efficiency fall into this category, e.g, pruning~\cite{jiang2022model} and using quantized models~\cite{reisizadeh2020fedpaq}, which are also helpful for reducing computation.


A small subset of the proposed approaches specifically devote their attention to resolving the computational issues in FL. In~\cite{xu2021helios}, a ``soft-training'' method was introduced to dynamically compress the original training model into a smaller restricted volume through rotating parameter training. In each round, it lets different parts of model parameters alternately join the training, but maintains the complete model for federated aggregation. The authors of~\cite{wang2021progfed} suggested dividing the model into sub-models, then using only a few sub-models for partial federated training while keeping the rest of the parameters fixed. During training, sub-model capacities are gradually increased until it reaches the full
model. Along similar lines, federated dropout~\cite{caldas2018expanding} is a technique that enables each client to locally operate on a smaller sub-model while still providing updates that can be applied to the larger global model on the server. Finally, the technique presented in~\cite{thapa2020splitfed}, called splitfed learning, combines the strengths of FL and split learning by splitting a NN into client-side and server-side sub-networks during federated training.


Our federated hyperdimensional computing approach is orthogonal to the most of the existing communication-efficient FL methods. For instance, it can be used in tandem with compression, subsampling, and client selection or techniques that reduce model complexity. In fact, in Section~\ref{sec:strategies}, we include some strategies for further improving the communication cost leveraging the statistical properties of hypervectors, even though HD models are much smaller (around hundred thousand parameters vs millions/billions), thus more communication-efficient, compared to DNNs. Furthermore, different from the aforementioned works, we account for unreliable communication scenarios. We use the robustness of HDC to tolerate communication errors and carry out accurate training. Finally, there are studies that aim at making the compute intensive DNN-based FL methods more efficient as summarized above. In contrast, HDC itself is a very light-weight framework with low computational cost. It was shown in previous work that HDC provides 3$\times$ reduction in training time, 1.8$\times$ in energy comsumption compared to optimized DNNs on NVIDIA Jetson TX2 low-power edge GPU~\cite{khaleghi2022generic}. An ASIC implementation of HDC for edge devices further improves the energy consumption by 1257$\times$ and training time by 11$\times$ over DNNs.

\section{Hyperdimensional Computing}

In the following, we first introduce and give an overview on hyperdimensional computing. We next analyze  hyperdimensional computing classification algorithm, then express it in a standard mathematical framework from statistical learning and optimization. The goal of this section is to provide an in-depth formal treatment of HDC as a general `learning' method. Leveraging the analysis presented here, we later study the convergence properties of federated hyperdimensional computing in Section \ref{sec:convergeanalysis},


\subsection{Background}
\label{sec:hdbackground}

HDC performs cognitive tasks using high-dimensional vectors, also known as hypervectors. The typical length of a hypervector is usually from 1,000 to 10,000 dimensions. Hypervectors are random with independent identically distributed (i.i.d.) components. THus, any randomly chosen pair of points in the hyperdimensional space are nearly-orthogonal~\cite{kanerva2000random}. 

\begin{figure*}[ht]
\centering
\subfloat[HDC encoding functionality]{\includegraphics[height=0.275\textwidth, keepaspectratio]{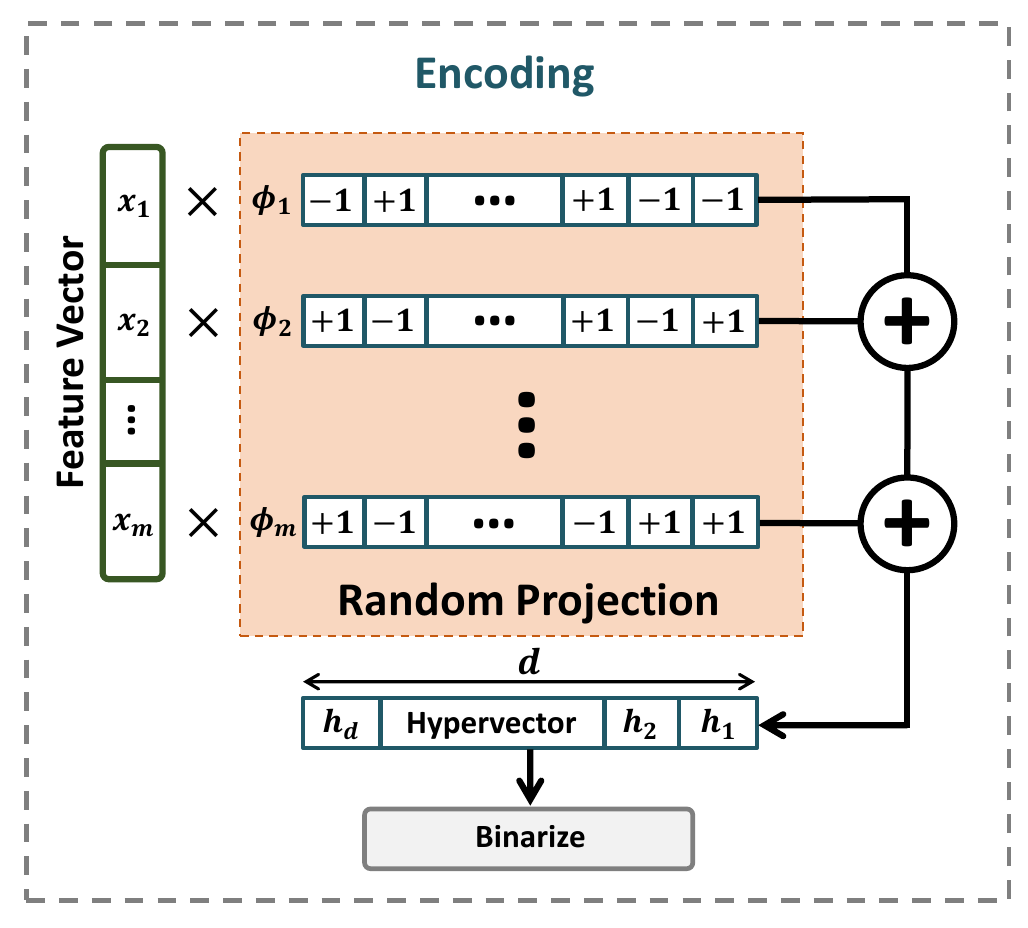}\label{fig:hdcbackground1}}
\subfloat[HDC classification overview]{\includegraphics[height=0.275\textwidth, keepaspectratio]{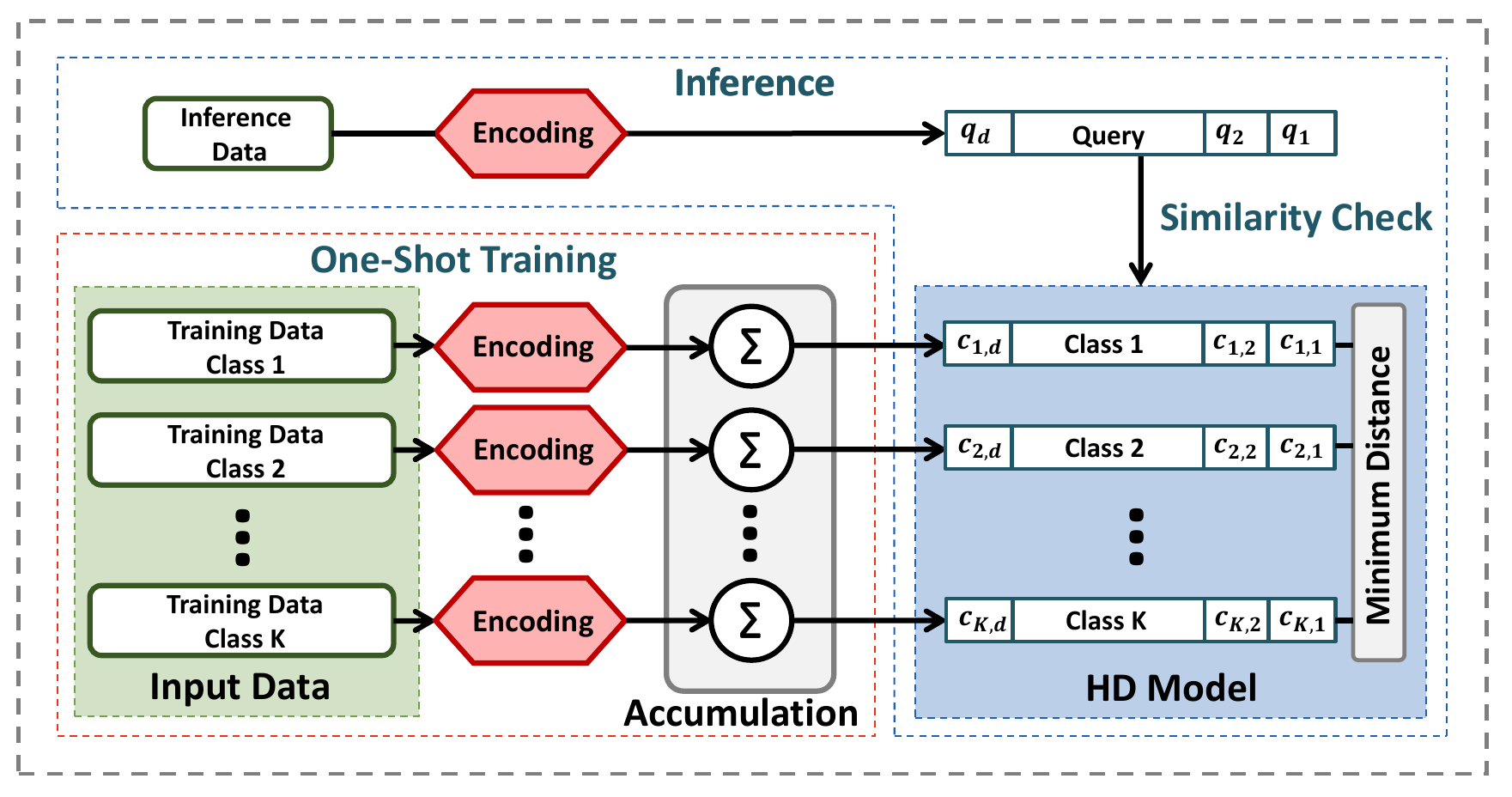}\label{fig:hdcbackground2}}
\caption{Hyperdimensional learning overview}
    \label{fig:hdcbackground}
\end{figure*}

The first step of HDC is to map/encode the input signal (e.g., an image, feature vector, or a time-series window) into
hypervectors. This step is common between all HDC applications. Assume the input $\mathbf{x} \in \mathcal{X}$ is represented by the vector $\mathbf{x} = [x_{1}, x_{2}, ..., x_{m}]^{T}$ where $x_{i}$s denote the features and $m$ is the length of input vector. HDC encoding operation maps the input data to its high-dimensional representation $\mathbf{h} \in \mathcal{H}$ with dimension $d\gg m$ under some function $\phi : \mathcal{X} \rightarrow \mathcal{H}$. There are
several encoding algorithms in literature with different memory-compute trade-offs, namely, base-level (a.k.a position-ID)~\cite{imani2017voicehd}, permutation~\cite{rahimi2016robust}, and random projection~\cite{imani2019bric}. In this work, we refer to the random projection encoding, but our methodology can be extended to any other encoding approach. Random projection encoding embeds the data into a high-dimensional Euclidean space under a random linear map. The output of this mapping can be quantized with minimal loss of information for better computational efficiency. If quantized,  the HD embedding is constructed as $\phi(\mathbf{x}) = \text{sign}(\mathbf{\Phi}\mathbf{x})$ under the encoding function $\phi : \mathbb{R}^{m} \rightarrow \mathbb{Z}^{d}$, the rows of which $\mathbf{\Phi} \in \mathbb{R}^{d \times m}$ are generated by randomly sampling directions from the $m$-dimensional unit sphere. Here, $\text{sign}(\mathbf{\Phi}\mathbf{x})$ is the element-wise sign function returning $+1$ if $\mathbf{\Phi}\mathbf{x} \geq 0$ and $-1$ otherwise. Fig.~\ref{fig:hdcbackground}a shows an overview of HDC encoding.


\subsection{Hyperdimensional Learning}
\label{sec:hyperdimensionallearning}




Many learning tasks can be implemented in the HD
domain. Here, we focus on classification, one of the most popular supervised learning problems. Suppose we are given collection of labeled examples $\mathcal{D} = \{(\mathbf{x}_{i},y_{i})\}^{n}_{i=1}$ where $\mathbf{x}_{i} \in \mathcal{X} \subset \mathbb{R}^{m}$ and $y_{i} \in \mathcal{C}$ is a categorical variable indicating the class label of a particular data sample $\mathbf{x}_{i}$. For HD learning, we first encode the entire set of data samples in $\mathcal{D}$ into hyperdimensional vectors such that $\mathbf{h}_{i} = \phi (\mathbf{x}_{i})$ is a hypervector in the \textit{d}-dimensional inner-product space $\mathcal{H}$. These high-dimensional embeddings represent data in a way that admits linear learning algorithms, even if the data was not separable to begin with. In other words, simple linear methods applied on HD encoded data can capture nonlinear decision boundaries on the original data~\cite{thomas2021theoretical}.

The common approach to learning with HD representations is to \textit{bundle} together the training
examples corresponding to each class into a set of ``prototypes'', which are then used for classification. The bundling operator is used to compile a set of elements in $\mathcal{H}$ and assumes the form of a function $\oplus : \mathcal{H} \times \mathcal{H} \rightarrow \mathcal{H}$. The function takes two points in $\mathcal{H}$ and returns a third point similar to both operands. We bundle all the encoded hypervectors that belong to the k-th class to construct the corresponding prototype $\mathbf{c}_{k}$:
\begin{equation}
    \mathbf{c}_{k} = \underset{i \text{ s.t. } y_{i} = k}{\bigoplus} \mathbf{h}_{i}
    \label{eq:bundling}
\end{equation}
Given a query data $\mathbf{x}_{q} \in \mathcal{X}$ for which we search for the correct label to classify, we take the encoded hypervector $\mathbf{h}_{q} \in \mathcal{H}$ and return the label of the most similar prototype:
\begin{equation}
    \hat{y}_{q} = k^{*} = \underset{k \in 1,...,K}{\text{argmax }}\delta(\mathbf{h}_{q},\mathbf{c}_{k})
    \label{eq:similaritycheck}
\end{equation}
where $\delta$ is a similarity metric.

\textit{One-Shot Training.} The bundling operator $\oplus$ is often chosen to be element-wise sum. In this case, the class prototypes are obtained by adding all hypervectors with the same class label. Then, the operation in Equation (\ref{eq:bundling}) is simply calculated as:
\begin{equation}
    \label{eq:oneshotraining}
    \mathbf{c}_{k} = \sum_{i \text{ s.t. } y_{i} = k} \mathbf{h}_{i}
\end{equation}
This can be regarded as a single pass training  since the entire dataset is only used once---with no iterations---to train the model (class prototypes). 

\textit{Inference.} The similarity metric $\delta$ is typically taken to be the cosine similarity which is a measure of angle between two vectors from an inner product space. Equation (\ref{eq:similaritycheck}) is rewritten using a dot-product and a magnitude operation as follows under cosine similarity:
\begin{equation}
    \hat{y}_{q} = k^{*} =  \underset{k \in 1,...,K}{\text{argmax }} \frac{\langle \mathbf{c}_{k},\mathbf{h}_{q} \rangle}{\|\mathbf{c}_{k}\|}
    \label{eq:similaritycheck2}
\end{equation}

\textit{Retraining.} One-shot training often does not result in sufficient accuracy for complex tasks. A common approach is to fine-tune the class prototypes using a few iterations of retraining~\cite{rahimi2016robust,imani2017voicehd,kanerva2009hyperdimensional,khaleghi2022generic}. We use the perceptron algorithm \cite{} to update the class hypervectors for mistpredicted samples. The model is updated only if the query in (\ref{eq:similaritycheck2}) returns an incorrect label. Let $y_{q}=k$ and $\hat{y}_{q} = k'$ be the correct and mispredicted labels respectively. Then, the new class prototypes after the retraining iteration are:
\begin{align}
    \mathbf{c}_{k} &= \mathbf{c}_{k} + \alpha\mathbf{h}_{q} \nonumber \\
    \mathbf{c}_{k'} &= \mathbf{c}_{k'} - \alpha\mathbf{h}_{q}
\end{align}
where $\alpha$ is the HD learning rate, controlling the amount of change we make to the model during each iteration. Figure~\ref{fig:hdcbackground}b shows an overview of HDC for classification.

\subsection{Hyperdimensional Linear Discriminant}
\label{sec:lineardiscriminant}
The single pass training and dot-product based inference approach of the HD algorithm bears a strong resemblance to Fisher’s linear discriminant~\cite{fisher1936use}. Assume that each sample $\mathbf{x} \in \mathcal{X}$ belongs to a class with binary label $y \in \{-1,1\}$ for notational convenience. The
assumption of a binary classication task is primarily for clarity of exposition, and our results can
be extended to support multi-class problems via techniques such as ``one-versus-rest'' decision rules. Fisher's linear discriminant on HD space finds the line $z = \mathbf{w}^{T}\mathbf{h}$ that best separates the two classes. The goal is to select direction $\mathbf{w}$ so that after projecting along this direction, \begin{enumerate*}[label=(\roman*)]
    \item the separation between classes are high with their means as far away as possible from each other, and
    \item the scatter within the classes is as small as possible with low variance.
\end{enumerate*}
A criterion that quantifies the desired goal is the \textit{Rayleigh quotient}:
\begin{align}
\label{eq:rayleighquotient}
    &J(\mathbf{w}) = \frac{\mathbf{w}^{T}\mathbf{S}_{B}\mathbf{w}}{\mathbf{w}^{T}\mathbf{S}_{W}\mathbf{w}}\\
    &\mathbf{S}_{B} = (\boldsymbol\mu_{1}-\boldsymbol\mu_{-1})(\boldsymbol\mu_{1}-\boldsymbol\mu_{-1})^{T} \nonumber \\
   &\mathbf{S}_{W} = \boldsymbol\Sigma_{1}+\boldsymbol\Sigma_{-1} \nonumber
\end{align}
where $\boldsymbol\mu_{\pm 1}$ and $\boldsymbol\Sigma_{\pm 1}$ are the mean vector and the covariance matrix respectively. $\mathbf{S}_{B}$ is defined as the between-class scatter which measures the separation between class means, while $\mathbf{S}_{W}$ is the within-class scatter, measuring the variability inside the classes. Our goal is achieved by maximizing the Rayleigh quotient with respect to $\mathbf{w}$. The corresponding optimal projection direction is then given as
\begin{equation}
    \label{eq:optimalw}
    \mathbf{w}^{*} = (\boldsymbol\Sigma_{1}+\boldsymbol\Sigma_{-1})^{-1}(\boldsymbol\mu_{1}-\boldsymbol\mu_{-1})
\end{equation}
One can use Fisher's linear discriminant method as a classifier where the decision criterion is a threshold on the dot-product (projection):
\begin{align}
    \label{eq:fisherclassifier}
    z = (\boldsymbol\mu_{1}-\boldsymbol\mu_{-1})^{T}(\boldsymbol\Sigma_{1}+\boldsymbol\Sigma_{-1})^{-1}\mathbf{h}_{q} + T \left\{
                \begin{array}{ll}
                 >0, \; \hat{y}_{q} = \:\:\,\, 1 \\
                 <0, \; \hat{y}_{q} = -1
                \end{array}
              \right.
\end{align}
In HD computing, the procedure of one-shot training followed by inference, described by (\ref{eq:oneshotraining}) and (\ref{eq:similaritycheck2}), is equivalent to above decision criterion. For two classes, the ``similarity check'' step in (\ref{eq:similaritycheck2}) can be rewritten in the form of a decision function as follows:
\begin{align}
    \hat{y}_{q} = \left\{
                \begin{array}{ll}
                \:\:\,\,1, \; \text{if } \frac{\langle \mathbf{c}_{1},\mathbf{h}_{q} \rangle}{\|\mathbf{c}_{1}\|} > \frac{\langle \mathbf{c}_{-1},\mathbf{h}_{q} \rangle}{\|\mathbf{c}_{-1}\|} \\
                -1, \; \text{if } \frac{\langle \mathbf{c}_{1},\mathbf{h}_{q} \rangle}{\|\mathbf{c}_{1}\|} < \frac{\langle \mathbf{c}_{-1},\mathbf{h}_{q} \rangle}{\|\mathbf{c}_{-1}\|}
                \end{array}
              \right.
\end{align}
which can be further simplified as:
\begin{align}
    \label{eq:classifier1}
    \hat{y}_{q} = \left\{
                \begin{array}{ll}
                \:\:\,\,1, \; \text{if } (\frac{\mathbf{c}_{1}}{\|\mathbf{c}_{1}\|} - \frac{\mathbf{c}_{-1}}{\|\mathbf{c}_{-1}\|})^{T}\mathbf{h}_{q} > 0 \\
                -1, \; \text{if } (\frac{\mathbf{c}_{1}}{\|\mathbf{c}_{1}\|} - \frac{\mathbf{c}_{-1}}{\|\mathbf{c}_{-1}\|})^{T}\mathbf{h}_{q} < 0
                \end{array}
              \right.
\end{align}
Since the class prototypes are normalized sums of hypervectors with the same labels, they relate to the respective class means by a scalar multiplication, i.e, $\mathbf{c}_{\pm 1} = \frac{\|\mathbf{c}_{\pm 1}\|}{N_{\pm 1}} \boldsymbol\mu_{\pm 1}$. Here, $N_{\pm 1}$ denotes the total number of samples in classes. We obtain the below decision rule after plugging in $\boldsymbol\mu_{\pm 1}$ into (\ref{eq:classifier1}), then dividing both sides of the inequalities by $\frac{\|\mathbf{c}_{\pm 1}\|}{N_{\pm 1}}$.
\begin{align}
    \label{eq:classifier2}
    \hat{y}_{q} = \left\{
                \begin{array}{ll}
                \:\:\,\,1, \; \text{if } (\boldsymbol\mu_{1}-\boldsymbol\mu_{-1})^{T}\mathbf{h}_{q} > 0 \\
                -1, \; \text{if } (\boldsymbol\mu_{1}-\boldsymbol\mu_{-1})^{T}\mathbf{h}_{q} < 0
                \end{array}
              \right.
\end{align}
Note that this is the same classifier as in (\ref{eq:fisherclassifier}) for the special case when $\boldsymbol\Sigma_{1} = \boldsymbol\Sigma_{-1} = \boldsymbol\Sigma = \frac{1}{2} \mathbf{I}$. HD encoding maps data points to a hyperdimensional space such that different dimensions of the hypervectors are uncorrelated, i.e, $\Sigma_{ij} \approx 0, \; i\neq j$. Therefore, one-shot training followed by inference in HD computing is equivalent to applying Fisher's linear discriminant and classifying sample encoded hypervectors. The above result shows the HD algorithm explicitly optimizes the discrimination between the data points from different classes. We first project data via HD encoding such that it becomes linearly separable, then find a linear discriminant.

\begin{figure}
    \centering
    \includegraphics[width = 8cm]{"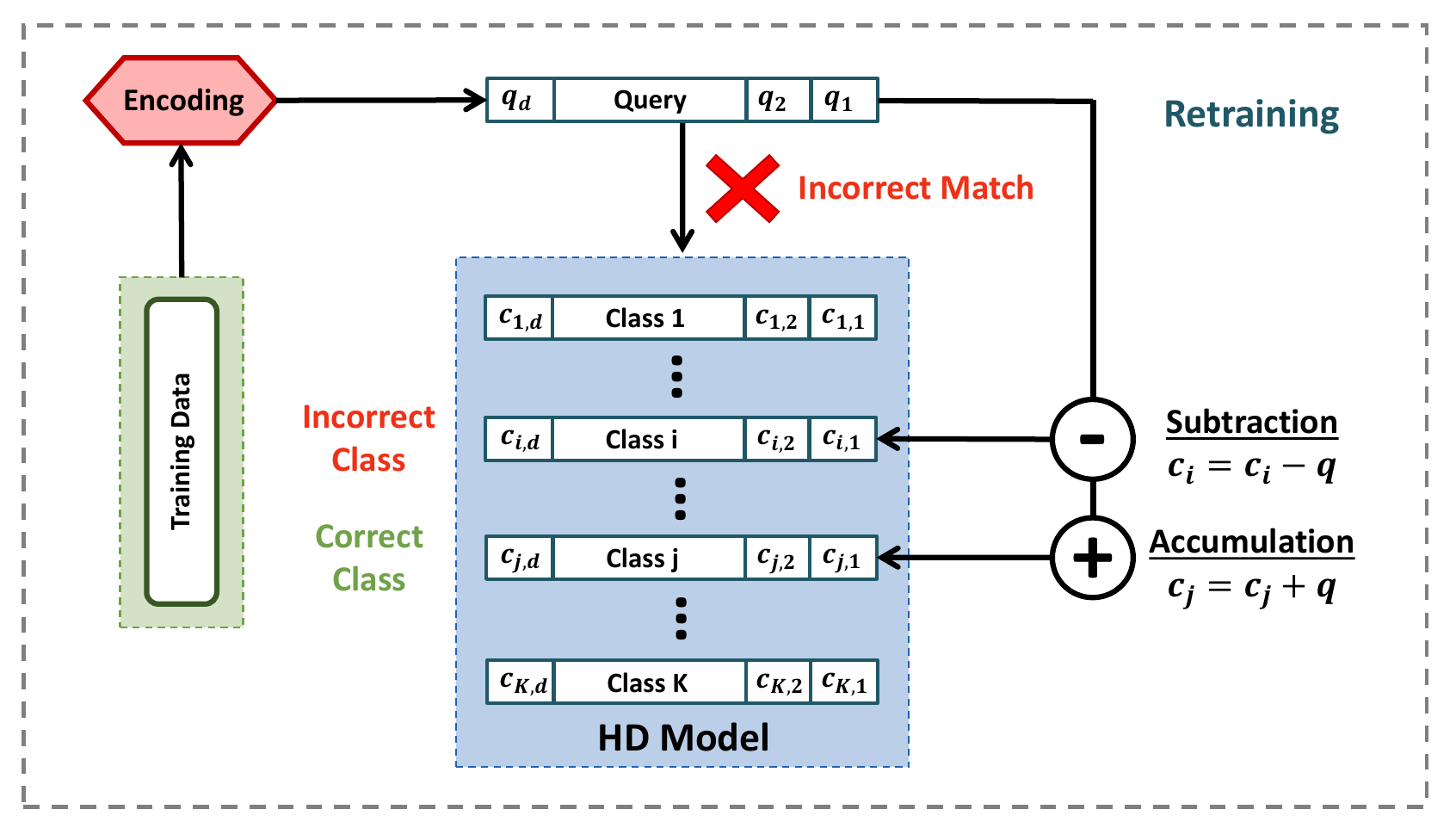"}
    \caption{HDC retraining}
    \label{fig:hdretraining}
\end{figure}

\subsection{A Gradient Descent Perspective on HDC}
\label{sec:hdgradientdescent}
A retraining step is required to fine-tune the HD model for tasks where one-shot training does not suffice. The goal is to update the class prototypes until finding the model that best separates classes. In the following, we analyze HD retraining process using the language of gradient methods from statistical learning and optimization. This viewpoint helps us provide a formal treatment of FedHDC as a general framework for federated learning, and precisely study its convergence properties.

Without loss of generality, we continue our analysis using binary class labels as in Section~\ref{sec:lineardiscriminant}. Let $\mathbf{w} \in \mathbb{R}^{d}$ be a vector of weights that specifies a hyperplane in the hyperdimensional space with $d$ dimensions. We define this vector in terms of class prototypes, such that $\mathbf{w} = \mathbf{c}_{1} - \mathbf{c}_{-1}$. Then, after inputing in the weight vector and simplifying the equations in (\ref{eq:classifier1}), classification of a query data $\mathbf{x}_{q}$ is made through the following decision function:
\begin{align}
    \label{eq:classifier3}
    \hat{y}_{q} = \left\{
                \begin{array}{ll}
                \:\:\,\,1, \; \text{if } \mathbf{w}^{T}\mathbf{h}_{q} > 0 \\
                -1, \; \text{if } \mathbf{w}^{T}\mathbf{h}_{q} < 0
                \end{array}
              \right.
\end{align}
This can be interpreted as a \textit{linear separator} on the HD representations of the data. It divides $\mathcal{H}$ into two half-planes, where the boundary is the plane with normal $\mathbf{w}$. The goal is to learn the weights such that all the positive examples ($y_{i} = 1$) are on one side of the hyperplane and all negative examples ($y_{i} = -1$) on the other. For the optimal set of weights, the linear function $g(\mathbf{h}) = \mathbf{w}^{T}\mathbf{h}$ agrees in the sign with the labels on all training instances, that is, $\text{sign(}\langle\mathbf{w},\mathbf{h}_{i}\rangle\text{)} = y_{i}$ for any $\mathbf{x}_{i} \in \mathcal{X}$. We can also express this condition as $y_{i} \langle \mathbf{w}, \mathbf{h}_{i} \rangle > 0$.

Recall that HD retraining, in the event of a misclassification, subtracts the query hypervector from the incorrect class prototype and adds it to the one that it should have been matched with. The two possible retraining iterations are illustrated below for binary classification.

\noindent\begin{minipage}{.5\linewidth}
\begin{align}
&\textbf{Misclassifying } \mathbf{x}_{1} \nonumber \\ 
&\mathbf{c}_{1} = \mathbf{c}_{1} + \alpha\mathbf{h}_{1} \nonumber \\
&\mathbf{c}_{-1} = \mathbf{c}_{-1} - \alpha\mathbf{h}_{1} \nonumber \\
\nonumber
\end{align}
\end{minipage}%
\begin{minipage}{.5\linewidth}
\begin{align}
&\textbf{Misclassifying } \mathbf{x}_{-1} \nonumber \\
&\mathbf{c}_{1} = \mathbf{c}_{1} - \alpha\mathbf{h}_{-1} \nonumber \\
&\mathbf{c}_{-1} = \mathbf{c}_{-1} + \alpha\mathbf{h}_{-1} \\
\nonumber
\end{align}
\end{minipage}
For both cases, the difference of class prototypes, i.e. $\mathbf{c}_{1} - \mathbf{c}_{-1}$, is updated as a function of the misclassified class label. A unified update equation that covers both cases is as follows:
\begin{equation}
    \mathbf{c}_{1} - \mathbf{c}_{-1} = \mathbf{c}_{1} - \mathbf{c}_{-1} + 2\alpha y_{i} \mathbf{h}_{i}
\end{equation}
Inputing in the weight vector in the above equation, we have:
\begin{equation}
    \mathbf{w} = \mathbf{w} + 2\alpha y_{i} \mathbf{h}_{i}
\end{equation}

\begin{algorithm}
\DontPrintSemicolon
  \KwInput{$\mathcal{D}$}
   set $t = 0, \mathbf{w}_{t} = 0$\;
  \Repeat{convergence}{\For{\text{random index} $i \in  \{1,...,n\} $}
    {
        \If{$y_{i} \langle \mathbf{w}_{t}, \mathbf{h}_{i} \rangle < 0$}
        {$\mathbf{w}_{t+1} = \mathbf{w}_{t} + \eta y_{i} \mathbf{h}_{i}$\;
     t = t + 1}
     }}
\caption{HD Retraining}
\end{algorithm}

A simple algorithm that implements HD retraining with the above notion of linear separators is described by Algorithm 1.
Here, $\eta$ is a positive scalar called the learning rate and $t$ denotes iteration number. We now show that
HD retraining can be represented as an instance of Empirical Risk Minimization (ERM). Particularly, we frame the retraining step as an optimization problem with convex loss function, then we argue that the updates in Algorithm 1 are equivalent to stochastic gradient descent (SGD) steps over an empirical risk objective. 

Our ultimate goal is to find the discriminant function $g_{\mathbf{w}}(\mathbf{h})$ which minimizes the empirical risk on the embedded training set $\mathcal{D}_{\mathcal{H}} = \{(\mathbf{h}_{1},y_{1}),...,(\mathbf{h}_{n},y_{n})\}$. Empirical risk is defined as follows:
\begin{equation}
\label{eq:empiricalrisk}
R_{emp}(g_{\mathbf{w}}) = \frac{1}{n}\sum_{i=1}^{n} \ell (g_{\mathbf{w}}(\mathbf{h}_{i}),y_{i})
\end{equation}
where $\ell : \mathcal{H} \times \mathcal{H} \rightarrow \mathbb{R}$ is a \textit{loss function} that describes the real-valued penalty calculated as a measure of the discrepancy between the predicted and true class labels. Zero empirical risk can be achieved if HD encoding admits a linearly separable representation. Otherwise, zero risk is not possible, but we search for the optimal weights that minimizes it:
\begin{equation}
    \mathbf{w}^{*} = \operatorname*{argmin}_{\mathbf{w}} R_{emp}(g_{\mathbf{w}})
\end{equation}

The ``no error'' condition, $y_{i} \langle \mathbf{w}, \mathbf{h}_{i} \rangle > 0 \; \forall i$, provides a very concise expression for the situation of zero empirical risk. It allows for the formulation of the learning problem as the following function optimization:
\begin{equation}
    \label{eq:optimization}
    \text{minimize} \;\; J(\mathbf{w}) = -\sum_{i=1}^{n} y_{i}\mathbf{w}^{T}\mathbf{h}_{i}
\end{equation}

The solution can be found by doing gradient descent on our cost function $J(\mathbf{w})$ where the gradient is computed as $\nabla J(\mathbf{w}) = -\sum_{i=1}^{n} y_{i}\mathbf{h}_{i}$. Another optimization method is the stochastic gradient descent that picks a random example at each step and makes an improvement to the model parameters. Then, the gradient associated with an individual example is $-y_{i}\mathbf{h}_{i}$. Given a loss function $\ell (\cdot)$, the stochastic gradient descent algorithm is defined below:

\noindent \textbf{Stochastic Gradient Descent:}

Given: starting point $\mathbf{w} = \mathbf{w}_{init}$, learning rates $\eta_{1},\eta_{2},\eta_{3},...$

\indent\indent(e.g. $\mathbf{w}_{init} = 0$ and $\eta_{t} = \eta$ for all $t$, or $\eta_{t} = 1/\sqrt{t}$).

For a sequence of random examples $(\mathbf{h}_{1},y_{1}),(\mathbf{h}_{2},y_{2}),...$
\begin{enumerate}
    \item Given example $(\mathbf{h}_{t},y_{t})$, compute the gradient $\nabla \ell (g_{\mathbf{w}}(\mathbf{h}_{t}),y_{t})$ of the loss w.r.t. the weights $\mathbf{w}$.
    \item Update: $\mathbf{w} \leftarrow \mathbf{w} - \eta_{t}\nabla \ell (g_{\mathbf{w}}(\mathbf{h}_{t}),y_{t})$
\end{enumerate}

To present an equivalent formulation to (\ref{eq:optimization}), consider the loss function $\ell (g_{\mathbf{w}}(\mathbf{h}),y) = \text{max}(0, y \langle \mathbf{w}, \mathbf{h} \rangle)$ for the empirical risk in (\ref{eq:empiricalrisk}). If $g_{\mathbf{w}}(\mathbf{h})$ has the correct sign, then we have a loss of 0, otherwise we have a
loss equal to the magnitude of $g_{\mathbf{w}}(\mathbf{h})$. In this case, if $g_{\mathbf{w}}(\mathbf{h})$ has the correct sign and is non-zero, then the gradient will be zero since an infinitesimal change in any of the weights
will not change the sign. So, the algorithm will not make any change on $\mathbf{w}$. On
the other hand, if $g_{\mathbf{w}}(\mathbf{h})$ has the wrong sign, then $\frac{\partial \ell}{\partial \mathbf{w}} = - y\mathbf{h}$. Hence, using $\eta_{t} = \eta$, the algorithm will update $\mathbf{w} \leftarrow \mathbf{w} + \eta y\mathbf{h}$. Note that this is exactly the same algorithm as HD retraining. We observe that empirical risk minimization by SGD with the above loss function gives us the update rule in Algorithm 1.

\section{FedHDC: Federated HD Computing}
\label{sec:fedhdc}
We study the federated learning task where an HD model is trained collaboratively by a loose federation of participating \textit{clients}, coordinated by a central \textit{server}. The general problem setting discussed in this paper mostly follows the standard federated averaging framework from the seminal work in~\cite{mcmahan2017communication}. In particular, we consider one central server and a fixed set of $N$ clients, each holding a local dataset. The $k$-th client, $k\in [N]$, stores embedded dataset $\mathcal{D}_{k} = \{(\mathbf{h}_{k,j},y_{k,j})\}^{n_{k}}_{j=1}$, with $n_{k} = |\mathcal{D}_{k}|$ denoting the number of feature-label tuples in the respective datasets.

The goal in FL is to learn a global model by leveraging the local data at the clients. The raw datasets cannot be shared with the central server due to privacy concerns, hence the training process is apportioned among the individual clients as described by the following distributed optimization problem:
\begin{equation}
\label{eq:globalobjective}
    \underset{\mathbf{w}}{\text{min }} \Big\{ F(\mathbf{w}) \triangleq \sum_{k=1}^{N} p_{k}F_{k}(\mathbf{w}) \Big\}
\end{equation}
where $p_{k}$ is the weight of the $k$-th client such that $p_{k} \geq 0$ and $\sum_{k=1}^{N} p_{k} = 1$. A natural and common approach is to pick $p_{k} = \frac{n_{k}}{n}$. Similar to Section~\ref{sec:hdgradientdescent}, we represent our HD model by a vector of parameters $\mathbf{w} \in \mathcal{H}\subseteq \mathbb{R}^{d}$. If the partition $\mathcal{D}_{k}$ is formed by randomly and uniformly distributing the training examples over the clients, then we have $\mathbb{E}_{\mathcal{D}_{k}}[F_{k}(\mathbf{w})] = F(\mathbf{w})$, where the expectation is over the set of examples assigned to the client. This is the IID assumption that usually does not hold in FL setting; $F_{k}$ could be an arbitrarily bad approximation to $F$ under non-IID data.

To define the learning objective and measure the fit of the model to data, we introduce a loss function as in (\ref{eq:empiricalrisk}). We denote $\ell \big(\mathbf{w} ; (\mathbf{h}_{k,j},y_{k,j}) \big)$ for the loss of the prediction on example $(\mathbf{h}_{k,j},y_{k,j})$ made with an HD model parametrized by $\mathbf{w}$. For the $k$-th client, the local objective $F_{k}(\cdot)$ is defined in the form of local empirical loss as follows:
\begin{equation}
\label{eq:localobjective}
    F_{k}(\mathbf{w}) = \frac{1}{n_{k}} \sum_{j=1}^{n_{k}} \ell \big(\mathbf{w} ; (\mathbf{h}_{k,j},y_{k,j}) \big)
\end{equation}
For ease of notation, we do not explicitly use $g_{\mathbf{w}}(\mathbf{h})$ to denote the learning model, instead substitute $\mathbf{w}$ which parametrizes it. The local empirical loss $F_{k}$ measures how well the client model fits the local data, whereas the global loss $F$ quantifies the fit to the entire dataset on average. We have shown above that the loss function $\ell  = \text{max}(0, y \langle \mathbf{w}, \mathbf{h} \rangle)$ captures the behavior of the HD algorithm for an equivalent optimization problem formulation solved by SGD. The objective is to find the model $\mathbf{w}^{*}$ that minimizes the global loss, i.e., $\mathbf{w}^{*} = \operatorname*{argmin}_{\mathbf{w}} F(\mathbf{w})$.

\begin{figure}
    \centering
    \includegraphics[width = 8.8cm]{"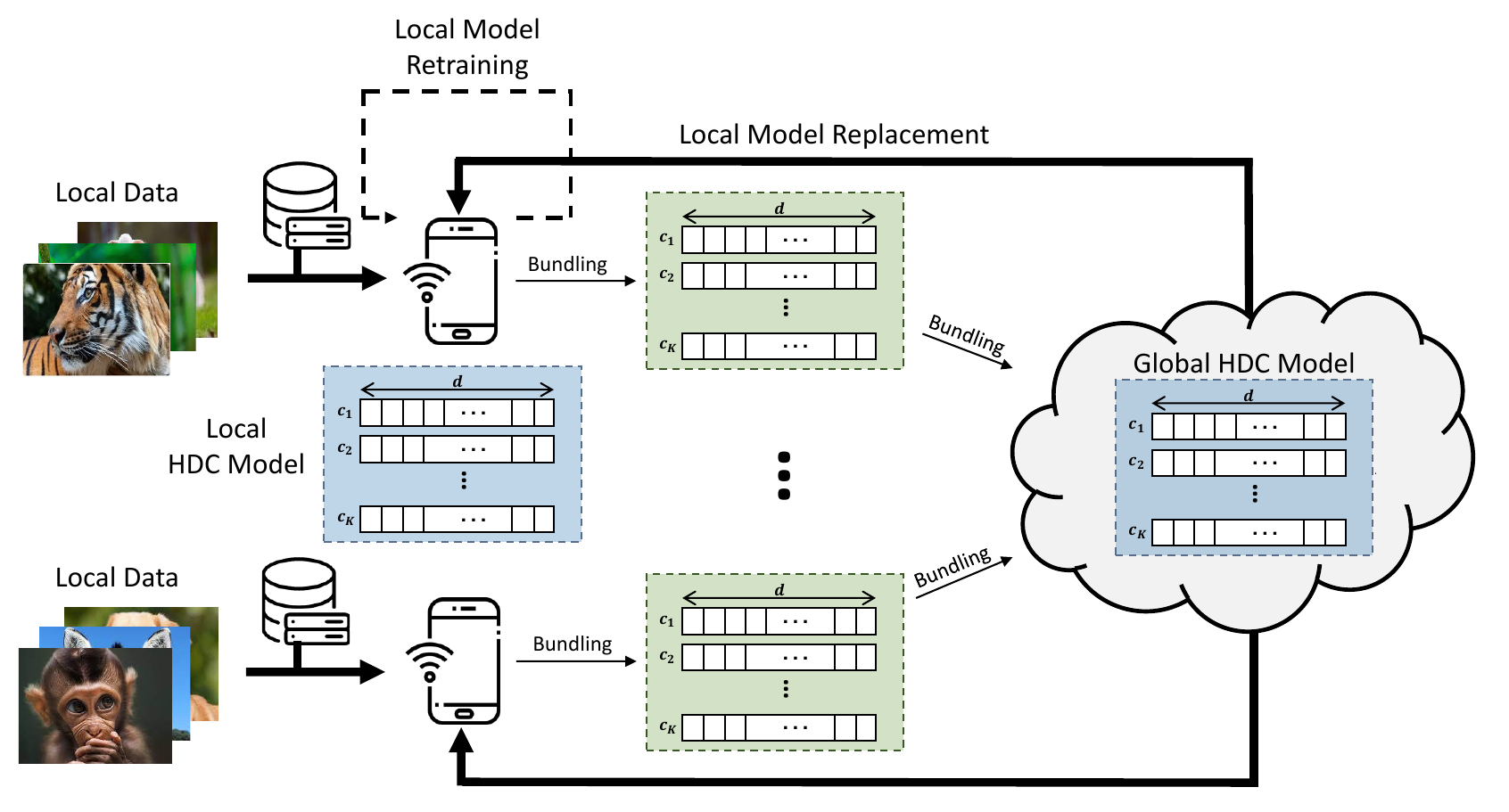"}
    \caption{FedHDC Training}
    \label{fig:fedhdctraining}
\end{figure}

\textbf{\textit{Algorithm.} }
In the \textit{federated bundling} framework, each client maintains its own HD model and participates in building a global model that solves (\ref{eq:globalobjective}) in a distributed fashion. This is achieved via an iterative training procedure for which we describe one round (say $t$-th) of the algorithm below.
\begin{enumerate}
    \item \textbf{Broadcast:} The central server broadcasts the latest global HD model, $\mathbf{w}_{t}$, to all clients.
    \item \textbf{Local updates:} Each client $k\in [N]$ sets its model $\mathbf{w}_{t}^{k} = \mathbf{w}_{t}$ and then performs training for $E$ epochs using local data:
    \begin{align}
        &\mathbf{w}_{t,0}^{k} = \mathbf{w}_{t}^{k}, \nonumber \\
        &\mathbf{w}_{t,\tau+1}^{k} \longleftarrow \mathbf{w}_{t,\tau}^{k}  - \eta_{t} \nabla F_{k}(\mathbf{w}_{t,\tau}^{k}, \xi_{\tau}^{k}),\; i=0,1,...,E-1, \nonumber \\
        &\mathbf{w}_{t+1}^{k} = \mathbf{w}_{t,E}^{k},
    \end{align}
    where $\eta_{t}$ is the learning rate and $\xi_{\tau}^{k}$ is a mini batch of data examples sampled uniformly from local dataset $\mathcal{D}_{k}$.
    \item \textbf{Aggregation:} The central server receives and aggregates the local models to produce a new global model:
    \begin{equation}
        \mathbf{w}_{t+1} = \sum_{k=1}^{N}p_{k}\mathbf{w}_{t+1}^{k}.
    \end{equation}
\end{enumerate}
After aggregation, the server moves on to the next round, $t+1$. This procedure is carried out until sufficient convergence is achieved. Fig.~\ref{fig:fedhdctraining} summarizes the federated training process for FedHDC. The overall update in one round of federated bundling is similar to a gradient descent step over the empirical loss corresponding to the entire distributed dataset across clients.

\subsection{FedHDC Convergence Analysis}
\label{sec:convergeanalysis}
In this section, we first specify the objective functions and the corresponding gradient computations in FedHDC, for whose general forms were discussed above. We then analyze the convergence behavior of FedHDC, showing that it converges to the global optimum at a rate of $\mathcal{O}(\frac{1}{T})$, where $T$ is the number of communication rounds.

For federated learning with HD algorithm, the optimization problem in (\ref{eq:globalobjective}) is cast as follows:
\begin{equation}
    \label{eq:globalobjectiveHD}
    \mathbf{w}^{*} = \operatorname*{argmin}_{\mathbf{w}} \sum_{k=1}^{N} \frac{p_{k}}{n_{k}} \sum_{j=1}^{n_{k}}  \text{max}(0, y_{j} \langle \mathbf{w}, \mathbf{h}_{j} \rangle),
\end{equation}
and the local gradient $\mathbf{g}_{k} = \nabla F_{k}(\mathbf{w})$ is computed at client $k\in [N]$ as:
\begin{equation}
    \label{eq:localgradient}
    \mathbf{g}_{k} = \frac{1}{n_{k}}\sum_{j=1}^{n_{k}} y_{j}\mathbf{h}_{j}
\end{equation}

As Equation (\ref{eq:localgradient}) suggests, the gradient computations are linear, demand low-complexity operations, and thus are favourable for resource-constrained, low-power client devices. However, in many learning tasks, linear federated learning models perform sub-optimally compared to their counterpart, DNN-based approaches. FedHDC diverges from traditional linear methods in this respect. It enjoys both the superior performance properties of non-linear models and low computational complexity of linear models. This is a direct result of HD computing, who embeds data into a high-dimensional space where the geometry is such that simple learning methods are effective. As we show in the following, linearity in HD training benefits convergence, at the same time the performance does not degrade due to the properties of non-linear hyperdimensional embeddings. Such convergence claims are not possible for non-convex and non-linear DNNs.

The functions $F_{k}(\cdot)$ and the gradients $\nabla F_{k}(\cdot)$ have the following properties:

\begin{enumerate}
    \item (L\textbf{-smoothness}). Each local function $F_{k}(\cdot)$ is L-smooth where the gradients $\nabla F_{k}(\cdot)$ are Lipschitz continuous: \textit{There exists a parameter $L>0$ such that for all $\mathbf{v}$,$\mathbf{w} \in \mathbb{R}^{d}$},
    \begin{equation}
        \| \nabla F_{k}(\mathbf{v}) - \nabla F_{k}(\mathbf{w})\| \leq L \| \mathbf{v} - \mathbf{w} \|. \nonumber
    \end{equation}
    \item (\textbf{Strong convexity}). Each local function $F_{k}(\cdot)$ is $\mu$-strongly convex and differentiable: \textit{For all $\mathbf{v}$,$\mathbf{w} \in \mathbb{R}^{d}$},
    \begin{equation}
        F_{k}(\mathbf{v}) \geq  F_{k}(\mathbf{w}) + (\mathbf{v} - \mathbf{w})^{T} \nabla F_{k}(\mathbf{w}) + \frac{\mu}{2}\| \mathbf{v} - \mathbf{w} \|^{2}. \nonumber
    \end{equation}
    \item (\textbf{Bounded variance}). The variance of stochastic gradients for each client $k$ is bounded: \textit{Let $\xi^{k}$ be sampled from the $k$-th client's dataset uniformly at random, then there exists constants $\sigma_{k}$ such that for all $\mathbf{w} \in \mathbb{R}^{d}$},
    \begin{equation}
        \mathbb{E} \| \nabla F_{k}(\mathbf{w},\xi^{k}) - \nabla F_{k}(\mathbf{w}) \|^{2} \leq \sigma_{k}^{2}. \nonumber
    \end{equation}
    
    \item (\textbf{Uniformly bounded gradient}). The expected squared norm of stochastic gradients is uniformly bounded: \textit{for all mini-batches $\xi^{k}$ at client $k\in [N]$ and for $\mathbf{w} \in \mathbb{R}^{d}$},
        \begin{equation}
        \mathbb{E} \| \nabla F_{k}(\mathbf{w},\xi^{k})\|^{2} \leq G^{2}. \nonumber
    \end{equation}
\end{enumerate}

\begin{figure*}[tp]
\begin{center}
  \includegraphics[width=18cm]{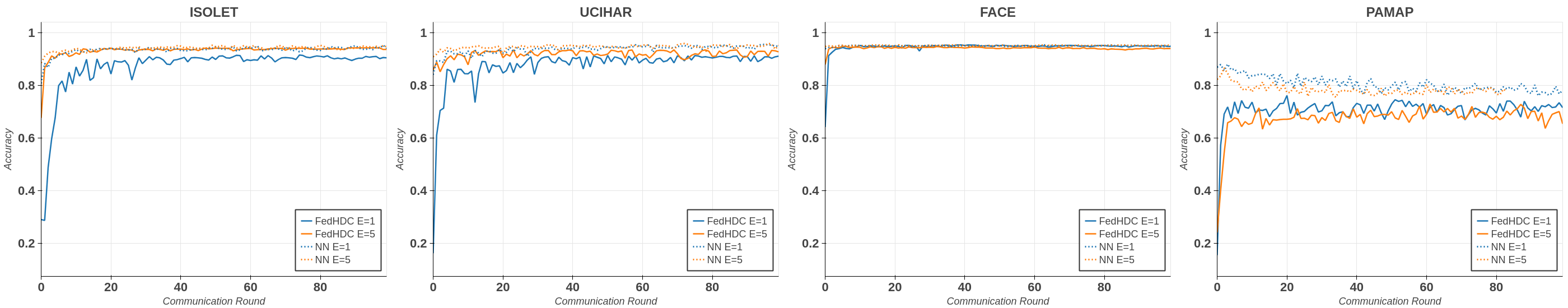}
  \caption{\small Accuracy and convergence of FedHDC and NN for various epochs (E)}
  \label{fig:figure1}
\end{center}
\end{figure*}

These conditions on local functions are typical and widely used for the convergence analysis of different federated averaging frameworks~\cite{li2019convergence,stich2018local}.

\textbf{Theorem 1.} \textit{Define $\kappa = \frac{L}{\mu}$, $\gamma = \text{max}\{8\kappa, E\}$ and choose learning rate $\eta_{t} = \frac{2}{\mu (\gamma + t)}$. Then, the convergence of FedHDC with Non-IID datasets and partial client participation satisfies}
\begin{equation}
    \mathbb{E}[F(\mathbf{w}_{T})] - F^{*} \leq \frac{2\kappa}{\gamma + T} \Big[\frac{B}{\mu} + \Big(2L+\frac{E\mu}{4}\Big) \| \mathbf{w}_{0} - \mathbf{w}^{*} \|^{2}\Big]
\end{equation}
\textit{where}
\begin{equation}
    B = \sum_{k=1}^{N} p_{k}^{2}\sigma_{k}^{2} + 6L\Gamma + 8(E-1)^{2}G^{2} + \frac{N-K}{N-1}\frac{4}{K}E^{2}H^{2}
\end{equation}
Here, $T$ is the number of communication rounds (or SGD steps), The term $\Gamma$ is used to quantify the degree of Non-IID~\cite{li2019convergence}. Let $F^{*}$ and $F^{*}_{k}$ be the minimum values of $F$ and $F_{k}$, respectively, then $\Gamma = F^{*} - \sum_{k=1}^{N}p_{k}F^{*}_{k}$. As shown in Theorem 1, FedHDC can achieve $\mathcal{O}(\frac{1}{T})$ convergence rate. Such claim does not hold for non-convex and non-linear DNNs. This result follows from the standard proof on the convergence of FedAvg on Non-IID data~\cite{li2019convergence}.  The proof is given in Appendix A.


\subsection{FedHDC Experiemental Results}


We implemented FedHDC on Python using a custom HDC library for the PyTorch framework. For FedHDC, we use hypervectors with dimension 10,000. For comparison, we use a NN with a fully connected layers with 128 units and
ReLU activation, and a final output layer with softmax.

To observe the performance of our approach focusing on the real-world use-cases, we evaluated FedHDC on a wide range of benchmarks shown in Table~\ref{tab:datasets} that range from relatively small datasets collected in a small IoT network to a large dataset that includes hundreds of thousands of face images. The data include: \textbf{ISOLET:} recognizing audio of the English alphabet, \textbf{UCIHAR:} detecting human activity
based on 3-axial linear acceleration and angular velocity
data, 
from different people, \textbf{PAMAP2:} classifying five human activities
based on a heart rate and inertial measurements, \textbf{FACE:} classifying images with faces/non-faces, and \textbf{MNIST:} recognizing handwritten digits by different people.



\begin{table}[h]
\centering
\label{tab:datasets}
\caption{Datasets ($n$: Feature Size, $K$: Number of Classes)}
\resizebox{!}{12.4mm}
{\begin{tabular}{c|c|c|c|c|c}
	\arrayrulecolor{black!60}\toprule
	     &   $n$  & $K$ &  \makecell{\textbf{Train}\\\textbf{Size}} & \makecell{\textbf{Test}\\\textbf{Size}} & \textbf{Description}\\
  \arrayrulecolor{black!60}\midrule
    \textbf{ISOLET} &  617 & 26 &  6,238 & 1,559 & Voice Recognition\\
    \textbf{UCIHAR} &  561 & 12 &  6,213 & 1,554 & Activity Recognition (Mobile)\\
    \textbf{PAMAP2} &  75 & 5 &  611,142 & 101,582 & Activity Recognition (IMU)\\
    \textbf{FACE} &  608 & 2 &  522,441 & 2,494 & Face Recognition \\
    \textbf{MNIST} &  784 & 10 &  60,000 & 10,000 & Handwritten Digit Recognition \\
     \arrayrulecolor{black!60}\bottomrule
\end{tabular}}
\end{table}


\subsubsection{Accuracy and Convergence}
We run our experiments for 100 clients and 100 rounds of communication. We first tune the hyperparameters for both FedHDC and CNNs, then experiment with different federated learning parameters. Fig.~\ref{fig:figure1} shows the accuracy and convergence of both FedHDC and the CNN for various number of local epochs $E$ and local batch sizes $B$. For all experiments, $C=0.2$ fraction of clients are randomly picked in every communication round. For all datasets, the best convergence is achieved with low number of epochs ($E=1$) and moderate batch sizes ($B=10,20$).

\begin{table}[h]
\centering
\label{tab:dimensionstudy}
\caption{Impact of Dimensionality on FedHDC Accuracy}
\begin{tabular}{c|c|c|c|c|c}
	\toprule
	\textbf{d} &   \textbf{1000} & \textbf{2000} & \textbf{4000} & \textbf{8000} & \textbf{10000}\\
  \arrayrulecolor{black!60}\midrule
    \textbf{ISOLET} & 90.79\% & 93.36\% & 95.07\% & 95.37\% & 94.59\% \\
    \textbf{UCIHAR} &  90.60\% & 93.98\% & 93.63\% & 93.54\% & 94.46\%\\
    \textbf{PAMAP2} &  74.9\% & 76.88\% & 76.10\% & 77.85\% & 77.98\% \\
    \textbf{FACE} &  95.05\% & 95.2\% & 95.74\% & 95.86\% & 96.17\% \\
    \textbf{MNIST} &  92.24\% & 93.81\% & 95.37\% & 96.34\% & 96.80\% \\
     \arrayrulecolor{black}\bottomrule
\end{tabular}
\end{table}

\subsubsection{Hypervector Dimensionality Study}
Table~\mbox{\ref{tab:dimensionstudy}} demonstrates the influence of hypervector dimensions on the FedHDC classification accuracy. A modest increase in accuracy is observed as the dimensionality grows. This outcome aligns with expectations, as the robustness of HDC is known to improve with increasing dimensions \mbox{\cite{thomas2021theoretical}}\mbox{\cite{rahimi2008random}} \mbox{\cite{kanerva1988sparse}}. Thomas \mbox{\cite{thomas2021theoretical}} showed that dimensionality is directly proportional to the bandwidth of the noise in HDC classification problems, thus providing a guideline for a tradeoff between noise and the hypervector size.  It is essential to consider the trade-off between performance and resource usage, as the computational cost rises with increasing dimensions.

In essence, the HD encoding dimension exhibits a linear relationship with the number of categorical features, while it depends logarithmically on the alphabet size. As previously mentioned, the separation quality of the problem is associated with factors such as the class separability and the encoding dimension. Intuitively, when the classes are well separated, a smaller encoding dimension can be employed to achieve satisfactory performance. This is because the inherent separability of the data aids in reducing the required dimensionality for efficient classification. Conversely, when the classes are poorly separated, a larger encoding dimension is necessary to enhance the robustness and accuracy of the classification process. Consequently, understanding the relationship between the HD encoding dimension and the problem's complexity is crucial for optimizing the performance of high-dimensional computing methods in various classification tasks.

\section{FHDnn: Federated Hyperdimensional Computing with CNN Feature Extraction}

\begin{table}[]
    \centering
    \caption{HDC on image data}
    \begin{tabular}{l||l|l|l|l|l}
    \hline
    Model       &   CIFAR10  &   CIFAR100   &   Flowers &   dtd &   GTSRB   \\ \hline\hline
    HD-linear   &   26.94   &   8.98    &   19.58   &   6.41    &   83.63    \\
    HD-non linear   &   41.98   &   20.35   &   25.68   &   8.13    &   84.11   \\
    HD-id level &   26.56   &   9.45    &   15.97 &  6.25    &   44.86   \\
    CNN & 90.1 & 78.4 & 81 & 98.7 & 94.6\\
    \hline
    \end{tabular}
    \label{tab:hd}
    \vspace{-0.5cm}
\end{table}

FedHDC gives great results for many datasets in a federated setting, but it does not have acceptable accuracy when doing complex image analysis due to inherent inaccuracy of HDC on larger images. Table \mbox{\ref{tab:hd}} summarizes accuracy of various state of the art encoding methods for HDC when running  image classification tasks \mbox{~\cite{dutta2022hdnn}}.  The current HD encoding methods are not able to match state of the art accuracy. In this section, to overcome this issue, we present FHDnn, a synergetic FL framework which combines CNNs and HDC. FHDnn uses a pre-trained CNN as a feature extractor, whose outputs are encoded into hypervectors and then used for training. It avoids the transmission of the CNN and instead trains only the HD learner in a federated manner. The CNN excels at learning a complex hierarchy of features and boasts high accuracy, whereas HDC provides efficient and robust training. Therefore, FHDnn enjoys the complimentary salient properties of both HDC and CNN to enable a lightweight, communication-efficient, and highly robust FL framework.

\begin{figure}
    \centering
    \includegraphics[width = 8.8cm]{"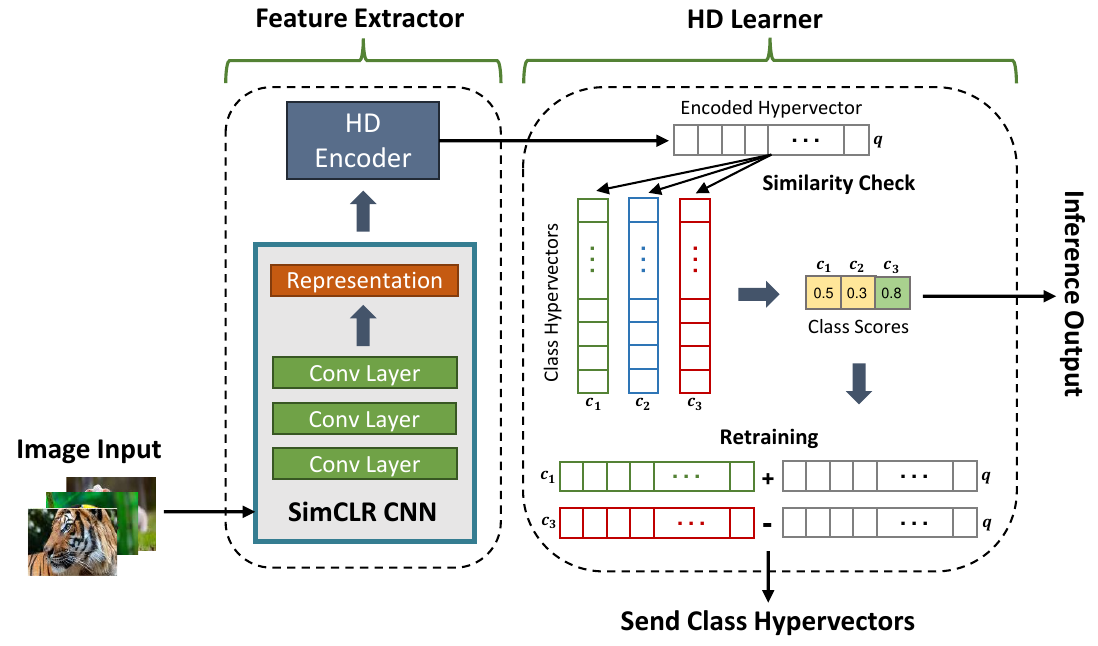"}
    \caption{FHDnn Model Architecture}
    \label{fig:architecture}
\end{figure}

\subsection{Model Architecture}
FHDnn consists of two components: i) a pre-trained CNN as a feature extractor and ii) a federated HD learner. Fig.~\ref{fig:architecture} shows the model architecture of FHDnn. The pre-trained feature extractor is trained once and not updated at run time. This removes the need for costly CNN weight updates via federated learning.   Instead, HD Computing is responsible for all the federated model updates.  Since its training only requires simple operations, it is much more efficient and scalable.  In the next subsections we describe both components.

\textbf{Feature Extractor:} While in theory any standard CNN can be used as a feature extractor, we use a pre-trained SimCLR ResNet model as our feature extractor due to its proven success in prior studies. SimCLR \cite{chen2020simple} is a contrastive learning framework which learns representations of images in a self-supervised manner by maximizing the similarity between latent space representations of different augmentations of a single image. This class-agnostic framework trained on a large image dataset allows for transfer learning over multiple datasets, (as evaluated in \cite{chen2020simple}) making it ideal for a generic feature extractor. Standard CNNs learn representations that are fine-tuned to optimize the classification performance of the dense classifier at the end of the network. Since SimCLR focuses on learning general representations as opposed to classification oriented representations, it is a better choice of a feature extractor. We choose the ResNet architecture due to availability of pre-trained models. It is possible to use other models such as MobileNet~\cite{howard2017mobilenets}.

\textbf{HD Learner:} FHDnn encodes the outputs of the feature extractor into hypervectors. More formally, given a point $\mathbf{x} \in \mathcal{X}$, the features $\mathbf{z} \subset \mathbb{Z}^n$ are extracted using the feature extractor $f: \mathcal{X} \rightarrow \mathbb{Z}$ where $f$ is a pre-trained neural network. The HD embedding is constructed as $\mathbf{h} = \phi(\mathbf{z}) = \text{sign}(\mathbf{\Phi}\mathbf{z})$ under the encoding function $\phi : \mathbb{Z} \rightarrow \mathcal{H}$. HD learner then operates on these hypervectors using binding and bundling which are simple and highly parallelizable. The goal of such configuration is to avoid the transmission of the CNN and instead train only the HD learner in a federated manner. An HD model is formed by bundling all encoded hypervectors with the same class level together. We perform bundling by the element-wise addition of those hypervectors, which generates corresponding class prototoypes. Then, the HD model is simply a set of hypervectors with the number of classes in the dataset. We use the HD learner in federated training that we discuss in the following.

\vspace{-1mm}
\subsection{Federated Training}
\begin{figure}
    \centering
    \includegraphics[width = 0.48\textwidth]{"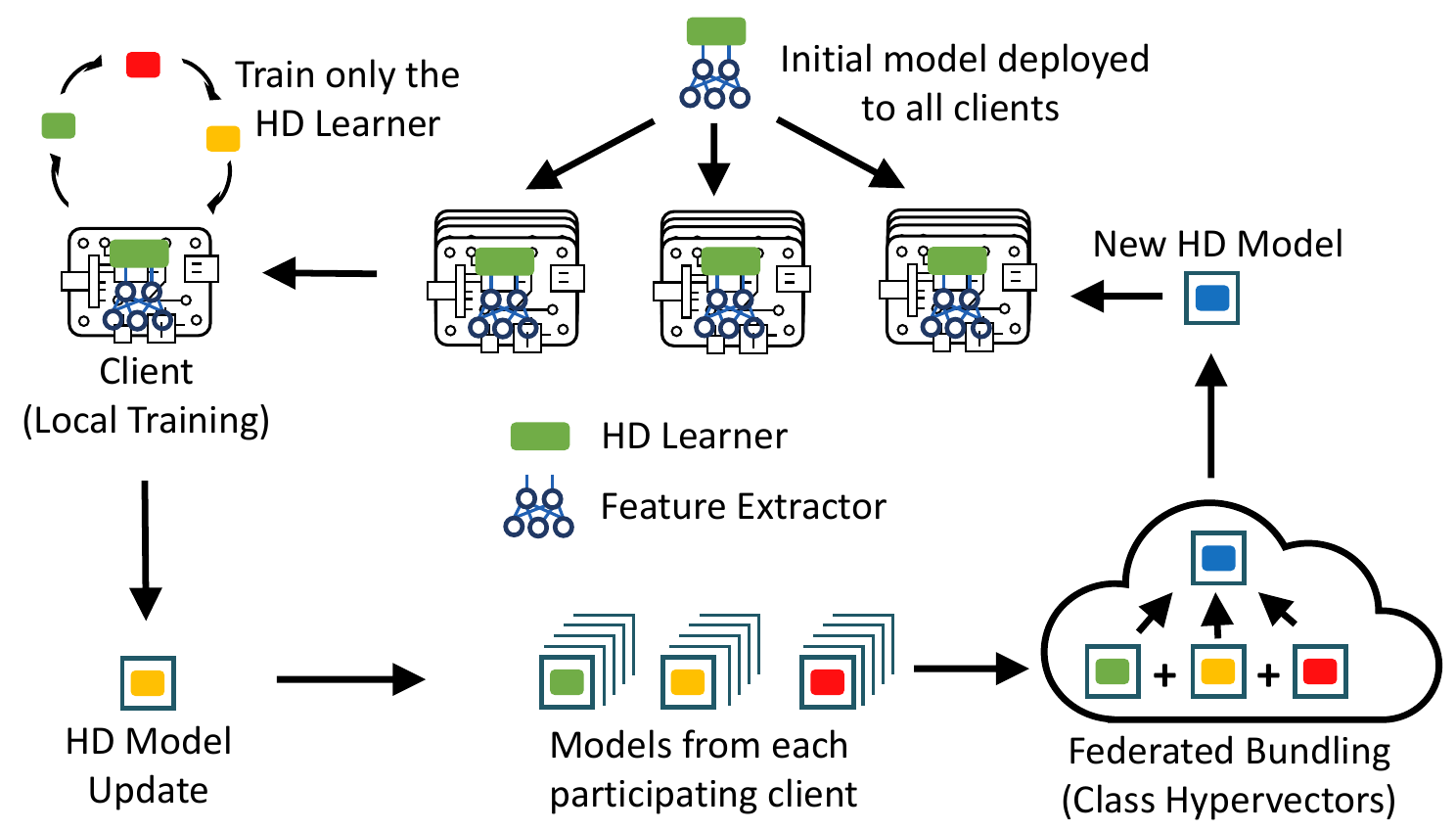"}
    \caption{FHDnn Federated Training}
    \label{fig:training}
\end{figure}

Fig.~\ref{fig:training} summarizes the overall federated training process for FHDnn. We separate the whole process into two steps, client local training and federated bundling. These two steps work in a cyclical fashion, one after the other, until convergence.

\textbf{Client Local Training:}
Each client initially starts the process with a feature extractor $f$ and an untrained HD learner. Once we get the encoded hypervectors using the method described above, we create class prototypes by bundling together hypervectors of the corresponding class using $\mathbf{c}_{k} = \sum_{i} \mathbf{h}_{i}^k$. Inference is done by computing the cosine similarity metric between a given encoded data point with each of the prototypes, returning the class which has maximum similarity.
After this one-shot learning process, we iteratively refine the class prototypes by subtracting the hypervectors from the mispredicted class prototype and adding it to the correct prototype as shown in Fig.~\ref{fig:architecture}. We define the complete HD model $\mathbf{C}$ as the concatenation of class hypervectors, i.e., $\mathbf{C} = [\mathbf{c}_{1}^{T},\mathbf{c}^{T}_{2},...,\mathbf{c}^{T}_{l}]$.

\textbf{Federated Bundling:}
In the federated bundling framework, each client maintains its own HD model and participates to build a global model in a distributed fashion. This is achieved via an iterative training procedure for which we describe one round (say $t$-th) of the algorithm below.
\begin{enumerate}[noitemsep, topsep=0pt, leftmargin=*]
    \item \textit{Broadcast:} The central server broadcasts the latest global HD model, $\mathbf{C}^{t}$, to all clients.
    \item \textit{Local updates:} Each participating client $k\in [N]$ sets its model $\mathbf{C}^{k}_{t} = \mathbf{C}_{t}$ and then performs training for $E$ epochs using local data.
    \item \textit{Aggregation:} The central server receives and aggregates the local models to produce a new global model:
    \begin{equation}
        \mathbf{C}_{t+1} = \sum_{k=1}^{N}\mathbf{C}_{t+1}^{k}.
    \end{equation}
\end{enumerate}
After aggregation, the server moves on to the next round, $t+1$. This procedure is carried out until sufficient convergence is achieved.

\subsection{FL Over Unreliable Channels With FHDnn}
\label{sec:unreliablecomm}
Federated learning is often carried out over wireless channels that attenuate the transmitted signal and introduce noise. Thus, the communication between clients and the server is unreliable, prone to transmission errors followed by packet losses. In this section, we show how FHDnn and FedHDC provide reliability for learning over unreliable wireless network at no overhead.  

We consider different models for the uplink and the downlink channels. The centralized server is assumed to be able to broadcast the models reliably, error-free at arbitrary rates, which is a common assumption in many recent works~\cite{li2019convergence, karimireddy2020scaffold,thapa2020splitfed,reisizadeh2020fedpaq,caldas2018expanding}. For uplink communications, the channel capacity per client is notably more constrained as the wireless medium is shared, so transmissions can be unreliable even at very low rates. We next describe the considered communication setup over such multiple access channels (MAC).


The mutual interference between the transmissions of multiple participating clients can lead to erroneous aggregation of models at the server. A common approach in FL to deal with interference is to use an orthogonal frequency division multiple access (OFDMA) technique~\cite{goldsmith2005wireless}. The resources of the shared-medium are partitioned in the time–frequency space and allocated among the clients. This way, each of the $N$ clients occupies one dedicated resource block, that is, channel's spectral band and time slot.

Even though each client model can be recovered separately due to the orthogonality, the distinct channels are still inherently noisy. The individual, independent uplink channels should be rate-limited to be treated as error-free links under the Shannon capacity theorem. 
However, the bandwidth allocated per client decreases with the number of clients, so does the capacity. Accordingly, the volume of data that can be conveyed reliably, i.e, throughput, scales by $1/N$. This implies that the data rates will be small, resulting in slow training speed unless transmission power is increased, which is undesirable considering energy consumption concerns.

Instead of limiting the rate to achieve error-free communication, we 
admit errors for the channel output at the server. The intuition is that the perturbations in the client models can be tolerated to a certain extent by the learning algorithm. If the learning model is robust to errors, then there is no need for forcing perfectly reliable transmissions. Thus, we analyze our FHDnn scheme assuming that the clients communicate over unreliable MAC and the transmitted models are corrupted.  




In the following, we consider three error models at different layers of the network stack. All models are applicable in practice depending on the underlying protocol. We first explore the properties of HD computing that makes the learning robust under the considered error models, then introduce different techniques for further improvement. 

\subsubsection{Noisy Aggregation}
In conventional systems, the transmitter performs three steps to generate the wireless signal from data: source coding, channel coding, and modulation. First, a source encoder removes the redundancies and compresses the data. Then, to protect the compressed bitsream against the impairments introduced by the channel, a channel code is applied. The coded bitstream is finally modulated with a modulation scheme which maps the bits to complex-valued samples (symbols), transmitted over the communication link.

The receiver inverts the above operations, but in the reverse
order. A demodulator first maps the received complex-valued channel output to a sequence of bits. This bitstream is then decoded with a channel decoder to obtain the original compressed data; however, it might be possibly corrupted due to the channel impairments. Lastly, the source decoder provides a (usually inexact) reconstruction of the transmitted data by applying a decompression algorithm.

For noisy aggregation, as an alternative of the conventional pipeline, we assume uncoded transmission~\cite{gastpar2008uncoded}. This scheme bypasses the transformation of the model to a sequence of bits, which are then need to be mapped again to complex-valued channel inputs. Instead, the real model parameter values are directly mapped to the complex-valued samples transmitted over the channel. Leveraging the properties of uncoded transmission, we can treat the channel as formulated in Equation (\ref{eq:channelequation}), where the additive noise is directly applied to model parameters. The channel output received by the server for client $k$ at round $t$ is given by
\begin{equation}
    \label{eq:channelequation}
    \mathbf{\tilde{w}}_{t}^{k} = \mathbf{w}_{t}^{k} + \mathbf{n}_{t}^{k}
\end{equation}
where $\mathbf{n}_{t}^{k} \sim \mathcal{N}(0,\, \sigma_{t,k}^{2})$ is the $d$-dimensional additive noise. The signal power and noise power are computed as $\mathbb{E} \| \mathbf{w}_{t}^{k}\|^{2} = P_{t,k}$ and $\mathbb{E} \| \mathbf{n}_{t}^{k}\|^{2} = \sigma_{t,k}^{2}$, respectively. Then, the signal-to-noise ratio (SNR) is:
\begin{equation}
    SNR_{t,k} = \frac{\mathbb{E} \| \mathbf{w}_{t}^{k}\|^{2}}{\mathbb{E} \| \mathbf{n}_{t}^{k}\|^{2} } = \frac{P_{t,k}}{\sigma_{t,k}^{2}}
\end{equation}

An immediate result of federated bundling is the improvement in the SNR for the global model. When the class hypervectors from different clients are bundled at the server, the signal power scales up quadratically with the number of clients $N$, whereas the noise power scales linearly. Assuming that the noise for each client is independent, we have the following relation:
\begin{equation}
\label{eq:overallsnr}
    SNR_{t} = \frac{\mathbb{E} \big[\sum_{k=1}^{N}\mathbf{w}_{t}^{k} \big]}{\mathbb{E} \big[\sum_{k=1}^{N}\mathbf{n}_{t}^{k}\big]} \approx \frac{N^{2}P_{t,k}}{N\sigma_{t,k}^{2}} = N \times SNR_{t,k}
\end{equation}
Notice that the effect of noise is suppressed by $N$ times due to bundling. This claim can also be made for the \mbox{FedAvg}~\cite{konevcny2016federated} framework over CNNs. However, even though the noise reduction factor is the same, the impact of the small noise might be amplified by large activations of CNN layers. In FHDnn, we do not have such problem as the inference and training operations are purely linear. 

One other difference of FHDnn from CNNs is its information dispersal property. HD encoding produces hypervectors which have holographic representations, meaning that the information content is spread over all the dimensions of the high-dimensional space. In fact, no dimension in a hypervector is more responsible for storing any piece of information than others. Since the noise in each dimension can be also assumed independent, we can leverage the information spread to further eliminate noise.

\begin{figure}
    \centering
    \includegraphics[width = 8cm]{"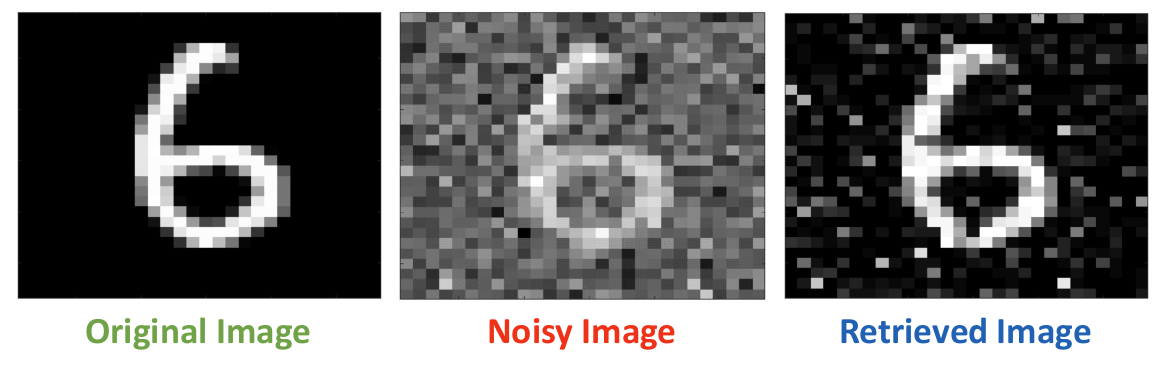"}
    \caption{Noise robustness of hyperdimensional encodings}
    \label{fig:mnistexample}
\end{figure}

Consider the random projection encoding described in Section~\ref{sec:hdbackground}, which is also illustrated by Fig.~\ref{fig:hdcbackground}a. Let the encoding matrix $\mathbf{\Phi} \in \mathbb{R}^{d \times n}$ expressed in terms of its $d$ row vectors, i.e., $\mathbf{\Phi} = [\Phi_{1}, \Phi_{2}, ..., \Phi_{d}]^{T}$. Then, the hypervector formed by encoding information $\mathbf{x} \in \mathcal{X}$ can be written as $\mathbf{h} = [\Phi_{1}^{T}\mathbf{x}, \Phi_{2}^{T}\mathbf{x},...,\Phi_{d}^{T}\mathbf{x}]^{T}$, where $\mathbf{x} = [x_{1}, x_{2}, ..., x_{n}]^{T}$. As implied by this expression, the information is dispersed over the hypervectors uniformly. Now consider additive noise over the same hypervector such that $\mathbf{h} + \mathbf{n} = [\Phi_{1}^{T}\mathbf{x} + n_{1}, \Phi_{2}^{T}\mathbf{x}+ n_{2},...,\Phi_{d}^{T}\mathbf{x}+n_{d}]^{T}$. We can reconstruct the encoded information from the noisy hypervector $\tilde{\mathbf{h}} = \mathbf{h} + \mathbf{n}$ as follows:
\begin{equation}
    \mathbf{x} \approx \Big[ \frac{1}{d} \sum_{i=1}^{d}\Phi_{i,1}\tilde{\mathbf{h}}_{i}, \frac{1}{d} \sum_{i=1}^{d}\Phi_{i,2}\tilde{\mathbf{h}}_{i}, ..., \frac{1}{d} \sum_{i=1}^{d}\Phi_{i,n}\tilde{\mathbf{h}}_{i} \Big]
\end{equation}
where $\tilde{\mathbf{h}}_{i} = \Phi_{i}^{T}\mathbf{x} + n_{i}$ are the elements of the noisy hypervector. The noise variance is then reduced by the averaging operation, similar to the case in Equation (\ref{eq:overallsnr}). Therefore, in HD computing, the noise is not only suppressed by bundling accross models from different clients, but also by averaging over the dimensions within the same hypervector. We demonstrate this over an example where we encode a sample from the MNIST dataset, add Gaussian noise, then reconstruct it. Fig.~\ref{fig:mnistexample} shows the original image, noisy image in the sample space, and reconstructed image for which the noise was added in the hyperdimensional space.

Finally, there is a ``flying under the radar'' principle for federated learning over noisy channel. The analysis in~\cite{wei2022federated} shows that since SGD is inherently a noisy process, as long as the channel noise do not dominate the SGD noise during model training, the convergence behavior is not affected. As the noise is immensely suppressed in FHDnn, we can claim such principle holds true in our case.

\subsubsection{Bit Errors}
\label{sec:biterrors}
We use bit error rate (BER) in conventional coded transmission as a figure of merit for system robustness. It is a measure on how accurately the receiver is able to decode transmitted data. The errors are bit flips in the received digital symbols, and are simply evaluated by the difference (usually Hamming distance) between the input bitstream of channel encoder and the output bitstream of channel decoder. Let $\hat{\mathbf{w}}$ be the binary coded model parameters that are communicated to the server. For the bit error model, we treat the channel as a binary symmetric channel (BSC), which independently flips each bit in $\hat{\mathbf{w}}$ with probability $p_{e}$ (e.g., $0 \rightarrow 1$). The received bitstream output at the server for client $k$ at round $t$ is then as follows:
\begin{equation}
    \label{eq:channelequation2}
    \mathbf{\tilde{\hat{w}}}_{t}^{k} = \mathbf{\hat{w}}_{t}^{k} \oplus \mathbf{e}_{t}^{k}
\end{equation}
where $\mathbf{e}_{t}^{k}$ is the binary error vector and $\oplus$ denotes modulo 2 addition. Given a specific vector $\mathbf{v}$ of Hamming weight $\text{wt}(\mathbf{v})$, the probability that $\mathbf{e}_{t}^{k} = \mathbf{v}$ is given by
\begin{equation}
    \mathbb{P}(\mathbf{e}_{t}^{k} = \mathbf{v}) = p_{e}^{\text{wt}(\mathbf{v})}(1-p_{e})^{m - \text{wt}(\mathbf{v})}
\end{equation}
The bit error probability, $p_{e}$, is a function of both the modulation scheme and the channel coding technique (assuming lossless source coding). To conclude the transmission, the corrupted bitstream in (\ref{eq:channelequation2}) is finally reconstructed to a real-valued model, i.e., $\mathbf{\tilde{\hat{w}}}_{t}^{k} \rightarrow \mathbf{\tilde{w}}_{t}^{k}$.

\begin{figure}
    \centering
    \includegraphics[width = 7.7cm]{"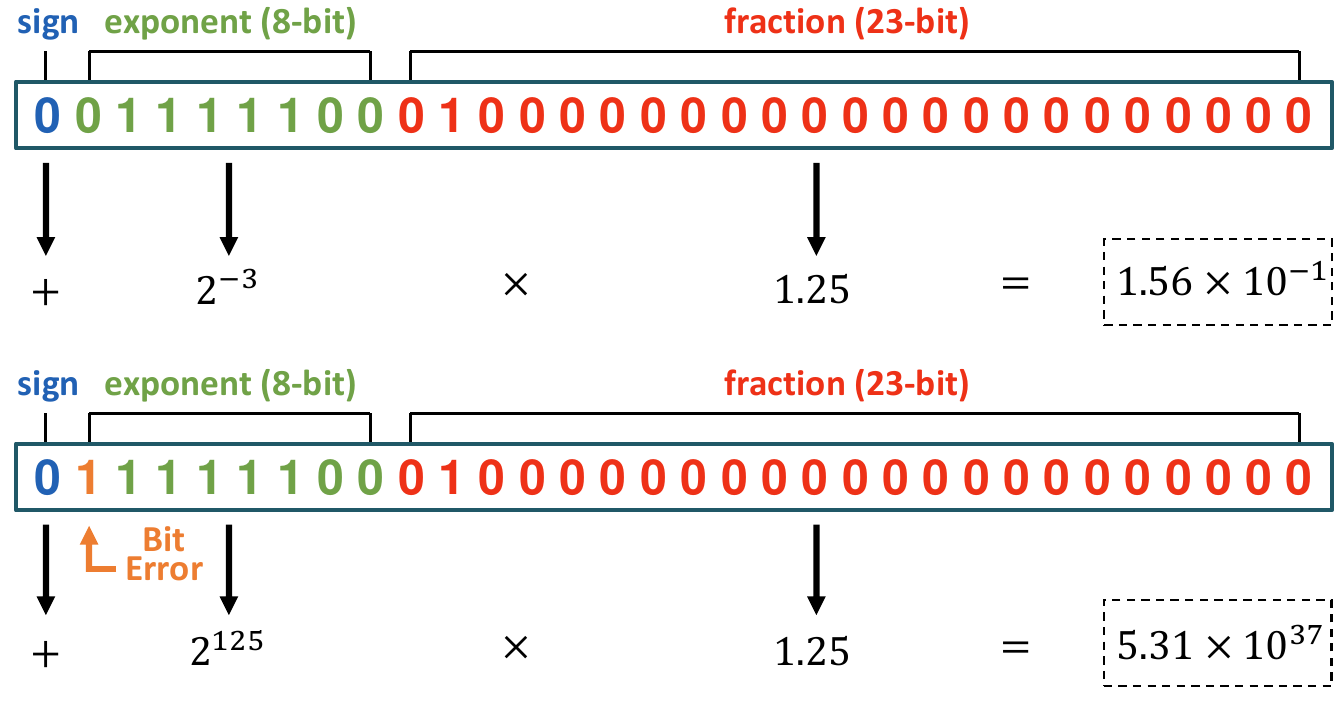"}
    \caption{Single bit error on a floating-point number}
    \label{fig:biterrorexample}
\end{figure}

Bit errors can have a detrimental effect on the training accuracy, especially for CNNs. At worst case, a single bit error in one client in one round can fail the whole training. In Fig.~\ref{fig:biterrorexample} we give an example of how much difference a single bit error can make for the standard 32 bit floating point CNN weights. In floating point notation, a number consists of three parts: a
sign bit, an exponent, and a fractional value. In \textit{IEEE 754} floating point representation, the sign bit is the most significant bit, bits 31 to 24 hold the exponent value, and the remaining bits contain the fractional value. The exponent bits represent a power of two ranging from -127 to 128. The fractional bits store a value between 1 and 2, which is multiplied by $2^{exp}$ to give the decimal value. Our example shows that one bit error in the exponent can change the weight value from $0.15625$ to $5.31\times 10^{37}$.


 The bit errors are contagious because a parameter from one client gets aggregated to the global model, then communicated back to all clients. Furthermore, errors propagate through all communication rounds because local training or aggregation does not completely change the parameter value, but only apply small decrements. For instance, assume a federated learning scenario with 100 clients and one bit error in a client's model as in the above example. After 10 rounds of training, the CNN weight for the global model will be on the order of $\sim \frac{5.31\times10^{37}}{100^{10}} = 5.31\times 10^{17}$, still completely failing the whole model. Consider ResNet-50, which has 20 million parameters, so training 100 clients even over a channel with $p_{e} = 10^{-9}$ BER results in two errors per round on average, making model failure inevitable.
 
  A similar problem exists with HD model parameters, but to a lesser extent because the hypervector encodings use integer representations. 
  Fig.~\ref{fig:biterrorexample2} implies that the parameters can also change significantly for the HD model. Particularly, errors in the most significant bits (MSB) of integer representation leads to higher accuracy drop. We propose a quantizer solution to prevent this.
  
\begin{figure}
    \centering
    \includegraphics[width = 8.5cm]{"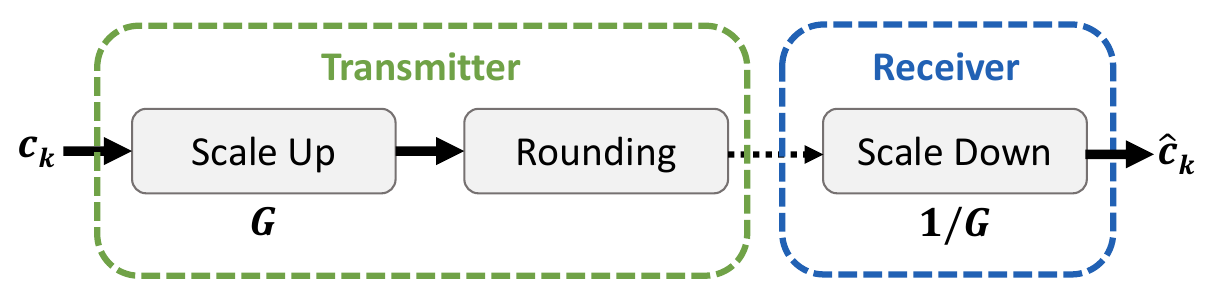"}
    \caption{Quantizer scheme}
    \label{fig:quantizer}
\end{figure}

\begin{figure}
    \centering
    \includegraphics[width = 8.8cm]{"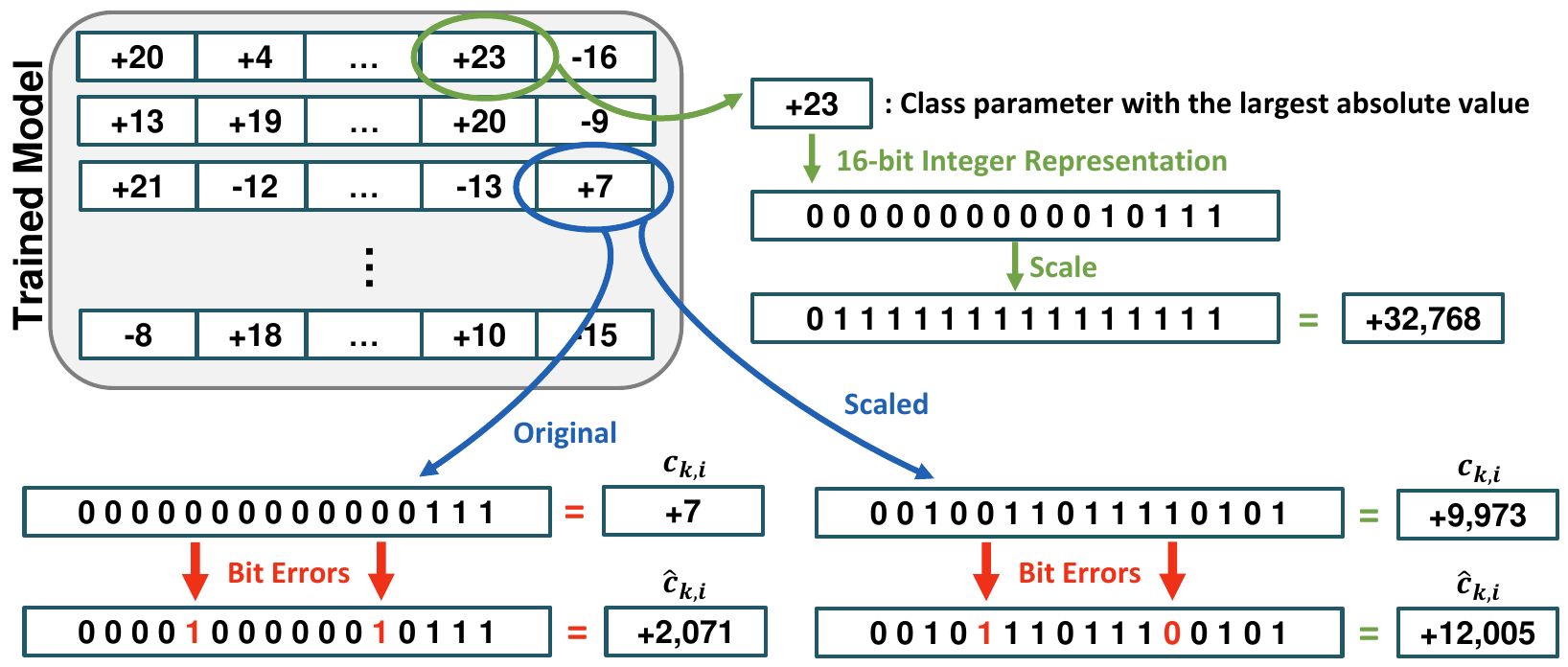"}
    \caption{An example scaling up operation}
    \label{fig:biterrorexample2}
\end{figure}

  The adopted quantizer design is illustrated in Fig.~\ref{fig:quantizer}. Inspired by the classical quantization methods in communication systems, we leverage \textit{scaling up} and \textit{scaling down} operations at the transmitter and the receiver respectively. This can be implemented by the automatic gain control (AGC) module in the wireless circuits. For a class hypervector $\mathbf{c}_{k}, \text{ } k\in \{1,...,K\}$, the quantizer output $Q(\mathbf{c}_{k})$ can be obtained via the following steps:
  \begin{enumerate}
      \item \textbf{Scale Up:} Each dimension in the class hypervector, i.e. $c_{k,i}$, is amplified with a scaling factor denoted quantization gain $G$. We adjust the gain such that the dimension with the largest absolute value attains the maximum value attainable by the integer representation. Thus, $G = \frac{2^{B-1}-1}{\max(c_{k})}$ where $B$ is the bitwidth.
      \item \textbf{Rounding:} The scaled up values are truncated to only retain their integer part.
      \item \textbf{Scale Down:} The receiver output is obtained by scaling down with the same factor $G$.
  \end{enumerate}

This way, bit errors are applied to the scaled up values. Intuitively, we limit the impact of the bit error on the models. Remember, from Equation (\ref{eq:similaritycheck2}), that prediction is realized by a normalized dot-product between the encoded query and class hypervectors. Therefore, the ratio between the original parameter and the received (corrupted) parameter determines the impact of the error on the dot-product. Without our quantizer, this ratio can be very large whereas after scaling \textit{up} then later \textit{down}, it is diminished. Fig.~\ref{fig:biterrorexample2} demonstrates this phenomenon. The ratio between the corrupted and the original parameter is $\frac{\hat{c}_{k,i}}{c_{k,i}} = \frac{2,071}{7} \approx 295.9$. The ratio decreases to only  $\frac{\hat{c}_{k,i}}{c_{k,i}} = \frac{12,005}{9,973} \approx 1.2$ between the scaled versions.


\subsubsection{Packet Loss}
At the physical layer of the network stack, errors are observed in the form of additive noise or bit flips directly on the transmitted data. On the other hand, at the network and transport layers, packet losses are introduced. The combination of network and protocol specifications allows us to describe the error characteristics, with which the data transmission process has to cope. 

The form of allowed errors, either bit errors or packet losses, are decided by the error control mechanism. For the previous error model, we assumed that the bit errors are admitted to propagate through the transport hierarchy. This assumption is valid for a family of protocols used in error resilient applications that can cope with such bit errors~\cite{wang2000error}. In some protocols, the reaction of the system to any number of bit errors is to drop the corrupted packets~\cite{kurose2005computer}. These protocols employ a cyclic redundancy check (CRC) or a checksum that allows the detection of bit errors. In such a case, the communication could assume bit-error free, but packet lossy link. We use the packet error rate (PER) metric as a performance measure, whose expectation is denoted packet error probability $p_{p}$. For a packet length of $N_{p}$ bits, this probability can be expressed as:
\begin{equation}
    \label{eq:packeterrorrate}
    p_{p} = 1 - (1-p_{e})^{N_{p}}
\end{equation}

The common solution for dealing with packet losses and guarantee successful delivery is to use a reliable transport layer
communication protocol, e.g., transmission control protocol (TCP), where various mechanisms including acknowledgment messages, retransmissions, and time-outs are employed. To detect and recover from
transmission failures, these mechanisms incur considerable communication
overhead. Therefore, for our setup we adopt user
datagram protocol (UDP), another widely used transport layer protocol. UDP is unreliable and cannot guarantee packet delivery, but is low-latency and have much less overhead compared to TCP.

HDC's information dispersal and holographic representation properties are also beneficial for packet losses. Another direct result of these concepts is obtaining partial information on data from any part of the encoded information. The intuition is that any portion of holographic coded information represents a blurred image of the entire data. Then, each transmitted symbol--packets in our case--contains an encoded image of the entire model.


\begin{figure}
\hspace{-1.5mm}\subfloat{\includegraphics[width=0.255\textwidth, keepaspectratio]{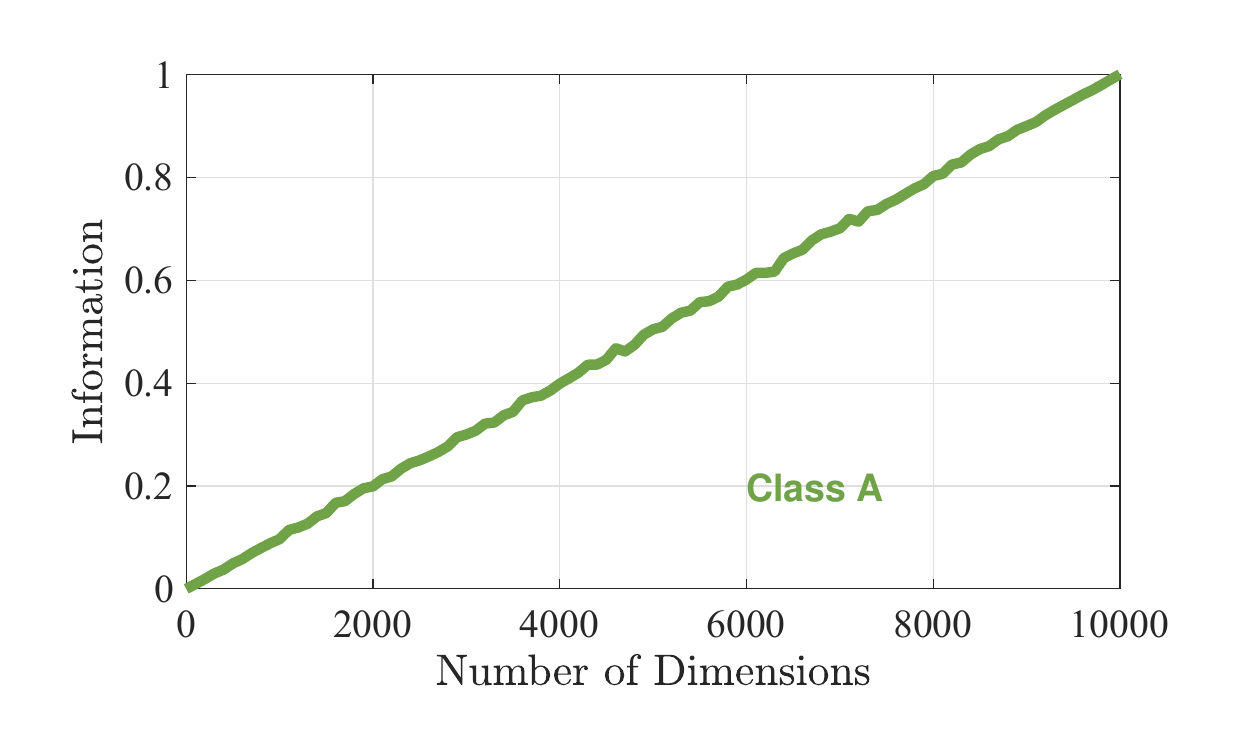}\label{fig:packetlossexample1}}\hspace{-3.6mm}
\subfloat{\includegraphics[width=0.255\textwidth, keepaspectratio]{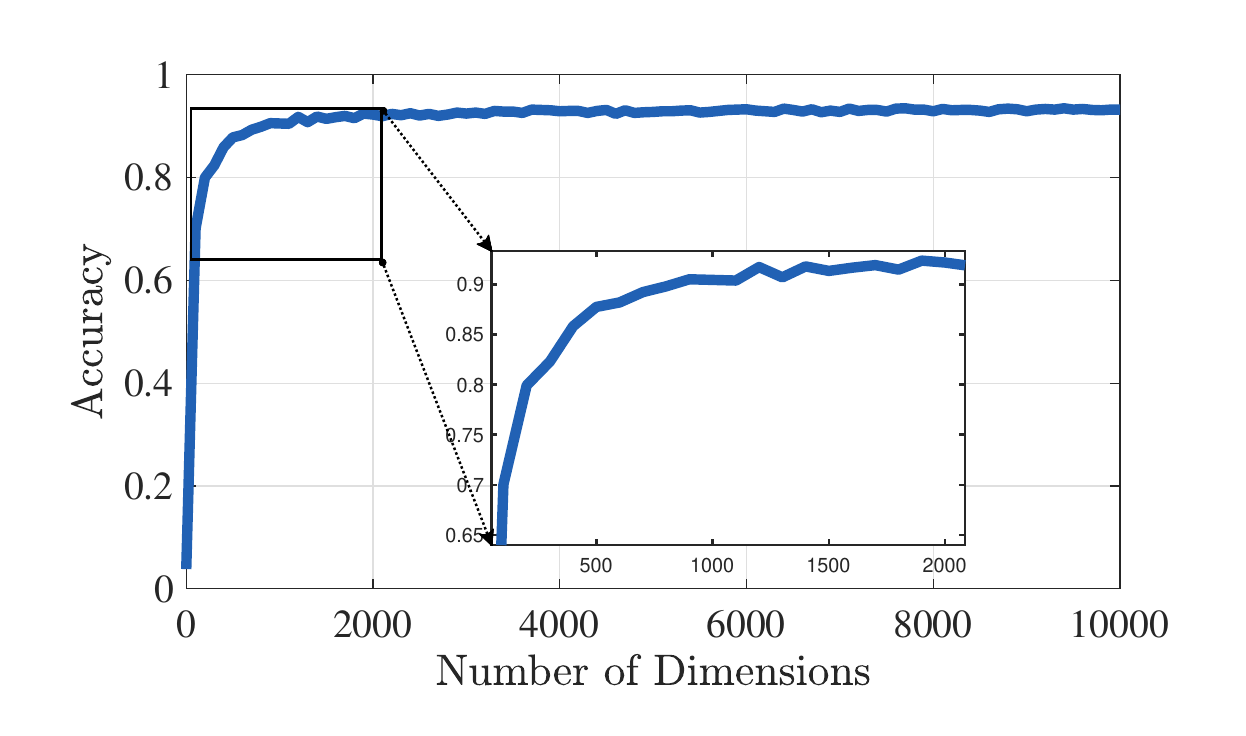}\label{fig:packetlossexample2}}
\caption{Impact of partial information on similarity check (left) and classification accuracy (right)}
    \label{fig:packetlossexample}
\end{figure}

We demonstrate the property of obtaining partial information as an example using a speech recognition dataset~\cite{isolet}. In Fig.~\ref{fig:packetlossexample}a, after training the model, we increasingly remove the dimensions of a certain class hypervector in a random fashion. Then we perform a similarity check to figure out what portion of the original dot-product value is retrieved. The same figure shows that the amount of information retained scales linearly with number of remaining dimensions. Fig.~\ref{fig:packetlossexample}b further clarifies our observation. We compare the dot-product values across all classes and find the class hypervector with the highest similarity. Only the relative dot-product values are important for classification. So, it is enough to have the highest dot-product value for the correct class, which holds true with $\sim90\%$ accuracy even when $80\%$ of the hypervector dimensions are removed.

\subsection{Strategies for Improving Communication Efficiency}
\label{sec:strategies}
The simplest implementation of FHDnn requires that clients send a full model back to the server in each round. Even though HDC models are much smaller than DNN models, it can still put a burden on communication. The structure and the characteristics of class hypervectors allow us to leverage certain techniques for improving communication efficiency of FHDnn. We propose three approaches: i) binarized differential transmission, ii) subsampling, and iii) sparsification \& compression.

\subsubsection{Binarized Differential Transmission} 
At the beginning of each round, the central server broadcasts the latest global HD model, $\mathbf{C}_{t}$, to all clients. Then, before performing local updates, each client makes a copy of this global model. Instead of sending the local updated models $\mathbf{C}_{t+1}^{k}$ at the aggregation step, the clients send the difference between the previous model and the updated model, i.e., $\mathbf{C}_{t+1}^{k}- \mathbf{C}_{t}$. We call this operation \textit{differential transmission}. As shown in (\ref{eq:binarize}), we binarize the difference to reduce the communication cost by 32x, going from 32-bit floating point to 1-bit binary transmission. 
\begin{equation}
\label{eq:binarize}
    \Delta \mathbf{C}_{bin}^{k} = \text{sign}(\mathbf{C}_{t+1}^{k}- \mathbf{C}_{t}), \; \forall k
\end{equation}
The central server receives and aggregates the differences, then adds it to the previous global model as:
\begin{equation}
    \mathbf{C}_{t+1} = \mathbf{C}_{t} + \sum_{k=1}^{N}\mathbf{C}_{bin}^{k}
\end{equation}
This global model is broadcasted back to the clients. Such binarization framework is not possible for the original federated bundling approach where clients communicate their full models. Binarizing the models itself instead of the `difference' results in unstable behavior in training. Therefore, we utilize binarized differential transmission whose stability can be backed by studies on similar techniques. In~\cite{bernstein2018signsgd}, it is theoretically shown that transmitting just the sign of each minibatch stochastic gradient can achieve full-precision SGD-level convergence rate in distributed optimization.

\subsubsection{Subsampling} 
In this approach, the clients only send a subsample of their local model to the central server. Each client forms and communicates a subsample matrix $\mathbf{\hat{C}}_{t+1}^{k}$, which is formed from a random subset of the values of $\mathbf{C}_{t+1}^{k}$. The server then receives and averages the subsampled client models, producing the global update $\mathbf{C}_{t+1}$ as:

\begin{equation}
    \mathbf{\hat{C}}_{t+1} = \frac{1}{N} \sum_{k=1}^{N}\mathbf{\hat{C}}^{k}_{t+1}
\end{equation}
The subsample selection is completely randomized and independent for each client in each round. Therefore, the average of the sampled models at the server is an unbiased estimator of their true average, i.e., $\mathbb{E} \|\mathbf{\hat{C}}_{t} \| = \mathbf{C}_{t}$. We can achieve the desired improvement in communication by changing the subsampling rate. For example, if we subsample 10\% of the values of $\mathbf{C}_{t+1}^{k}$, the communication cost is reduced by 10x.

\subsubsection{Sparsification \& Compression}
The goal of this approach is to drop the elements (class hypervector dimensions) of each individual class that have the
least impact on model performance. As discussed in Section~\ref{sec:hyperdimensionallearning}, given a query hypervector, inference is done by comparing it with all class hypervectors to find the one with the highest similarity. The similarity is typically taken to be the cosine similarity and calculated as a normalized dot-product between the query hypervector and class hypervectors. The elements of a query hypervector are input dependent and changes from one input to another one. Due to the randomness introduced by HDC encoding, the
query hypervectors, on average, have a uniform distribution of values in all dimensions. Under this assumption, we need to find and drop the elements of class hypervectors that have minimal impact on cosine similarity.  Indeed, the
elements with the smallest absolute values are the best candidates as they have the least contribution to the dot-product computation of cosine similarity.

We find the elements of each class hypervector with the smallest absolute value and make those elements zero. For example, for the $i^{th}$ class hypervector, we select $S$ elements with the minimum absolute value as follows.
\begin{equation}
    \text{min}\{c_{d}^{i},...,c_{2}^{i},c_{1}^{i}\}_{S}
\end{equation}
 To make a model with $S\%$ sparsity, we make $\frac{S}{100} \times d$ elements of each class hypervector zero. Then, we employ the Compressed Sparse Column (CSC)~\cite{lu2018spwa} to compress the sparse model. CSC stores only the non-zero data values and the number of zero elements between two consecutive non-zero elements.







\section{FHDnn Results}
We demonstrate through systematic experiments the performance of FHDnn under various settings. We first briefly discuss the datasets and setup for evaluation, and present our results for different data distributions under the reliable communication scenario. We then compare the resource usage of FHDnn against CNNs. The strategies for improving communication efficiency are also evaluated in this section. Lastly, we analyze FHDnn under three different unreliable network settings: packet loss, noise injection, and bit errors.



\begin{figure}
    \centering
    \includegraphics[width = 8.5cm]{"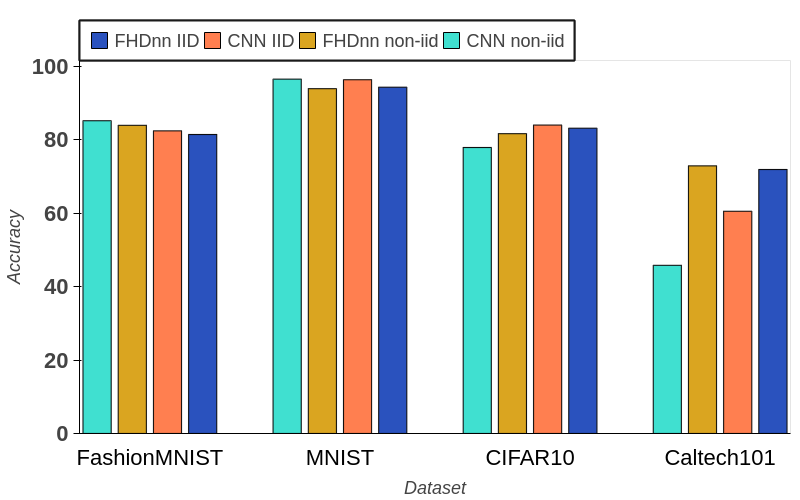"}
    \caption{Accuracy of FHDnn and ResNet on different datasets}
    \label{fig:acc}
\end{figure}

\subsection{Experimental Setup}
We evaluate FHDnn on 3 different real world datasets: MNIST\cite{deng2012mnist}, FashionMNIST\cite{fmnist}, CIFAR10 \cite{cifar} and Caltech101 \cite{li_andreeto_ranzato_perona_2022}. For the MNIST dataset, we use a CNN with two 5x5 convolution layers, two fully connected layers with 320 and 50 units and ReLU activation, and a final output layer with softmax. The first convolution layer has 10 channels while the second one has 20 channels, and both are followed by 2x2 max pooling. While for the CIFAR10 and FashionMNIST datasets, the well-known classifier model, ResNet-18 with batch normalization proposed in~\cite{he2016deep}, is used. We run our experiments on Raspberry Pi Model 3b and NVIDIA Jetson for the performance evaluations. All models are implemented on Python using the PyTorch framework. We consider an IoT network with $N=100$ clients and one server. The simulations were run for 100 rounds of communication each in order to keep our experiments tractable.


We first tune the hyperparameters of both FHDnn and CNNs, and analyze their performance by experimenting with three key parameters: $E$, the number of local training epochs, $B$ the local batch size, and $C$, the fraction of clients participating in each round. We select the best parameters for ResNet and use the same for FHDnn for all experiments in order to allow for a direct comparison. We study two ways of partitioning the datasets over clients: \textbf{IID}, where the data is shuffled and evenly partitioned into all clients, and \textbf{Non-IID}, where we first sort the data by their labels, divide it into a number shards of a particular size, and assign the shards to each of clients. 

We test FHDnn on two different types of edge devices: Raspberry PI 4 (RPi)\mbox{\cite{rpi}} and NVIDIA Jetson\mbox{\cite{jetson}}. The RPi features a Broadcom BCM2711 quad-core Cortex-A72 (ARM v8) 64-bit SoC, running at 1.5GHz, and 4GB RAM. The NVIDIA Jetson uses a quad-core ARM Cortex-A57 CPU, 128-core NVIDIA Maxwell GPU, and 4GB memory.

\begin{figure}
    \centering
    \includegraphics[width = \linewidth]{"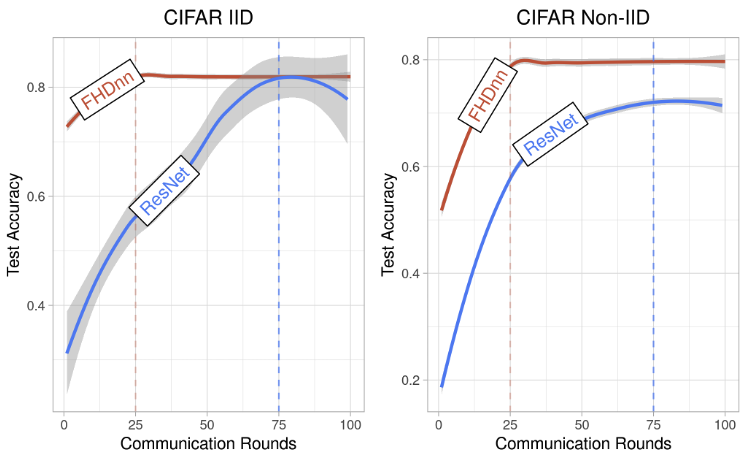"}
    \caption{Accuracy and Number of communication rounds for various hyperparameters}
    \label{fig:convergence}
\end{figure}

\begin{figure*}[]
    \centering
    \includegraphics[width = \textwidth]{"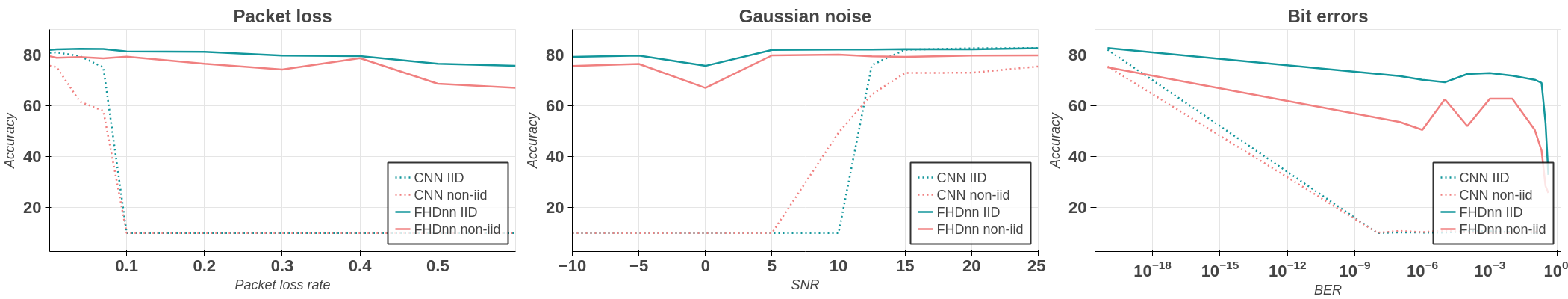"}
    \caption{Accuracy comparison of FHDnn with ResNet with noisy network conditions}
    \label{fig:loss}
\end{figure*}

\subsection{FHDnn Accuracy Results}\label{sec:rel_acc}
Fig.~\ref{fig:acc} compares the test accuracy of FHDnn with ResNet on MNIST, CIFAR-10, FashionMNIST and Caltech101 datasets after 100 rounds of federated training. We observe that FHDnn achieves accuracy comparable to the state of the art, even though it trains a much smaller and less complex model. We depict how test accuracy changes over communication rounds for CIFAR-10 in Fig.~\ref{fig:convergence}. The plot illustrates the smoothed conditional mean of test accuracy across all different hyperparameters (E,B,C) for IID and Non-IID distributions. FHDnn reaches an accuracy of 82\% in less than 25 rounds of communication whereas ResNet takes 75 rounds on average for both IID and Non-IID data distributions. Moreover the hyperparameters do not have a big influence for FHDnn as seen by the narrow spread (gray region) in Fig.~\ref{fig:convergence}. Note that the local batch size $B$ doesn't impact FHDnn at all due to the linear and additive nature of its training methodology. This allows us to use higher batch sizes up to the constraints of the device, allowing for faster processing, and going over the dataset in less rounds. On the other hand, the batch size $B$ affects the convergence of CNNs.

\subsection{FHDnn Performance and Energy Consumption}
Local training is computationally expensive for constrained IoT devices, which was one of the main drivers for centralized learning over many years. Particularly, CNN training involves complicated architectures and backpropagation operation that is very compute intensive. In addition, this has to be repeated for many communication rounds. HD on the contrary is lightweight, low-power, and fast. Table \ref{tab:time} quantitatively compares the computation time and energy consumption of FHDnn and ResNet local training on 2 different edge device platforms. FHDnn is $~35\%$ faster and energy efficient than ResNet on Raspberry Pi and $~80\%$ faster and energy efficient on the Nvidia Jetson.

\begin{table}[h]
\centering
\caption{Performance on Edge Devices}
\label{tab:time}
\begin{tabular}{|c|cc|cc|}
\hline
\multirow{2}{*}{\textbf{Device}} & \multicolumn{2}{c|}{\textbf{Training Time (Sec)}} & \multicolumn{2}{c|}{\textbf{Energy (J)}} \\ \cline{2-5} 
                  & \multicolumn{1}{c|}{FHDnn} & ResNet & \multicolumn{1}{c|}{FHDnn} & ResNet \\ \hline
Raspberry Pi      & \multicolumn{1}{c|}{\textbf{858.72}}  & 1328.04   & \multicolumn{1}{c|}{\textbf{4418.4}}      & 6742.8        \\ \hline
Nvidia Jetson     & \multicolumn{1}{c|}{\textbf{15.96}} & 90.55  & \multicolumn{1}{c|}{\textbf{96.17}}      & 497.572 \\ \hline
\end{tabular}
\end{table}

\subsection{FHDnn in Unreliable Communication}
In this section, we analyze the performance of FHDnn and ResNet under unreliable network conditions as described in Section \ref{sec:unreliablecomm}. We obtained similar results for FedHDC, which is why we focus on FHDnn results in this section.  Fig.~\ref{fig:loss} shows the performance of models under packet loss, Gaussian noise, and bit errors. To maintain a direct comparison between ResNet and FHDnn, we use the same hyperparameters for both models and all experiments. We set $E = 2, C = 0.2, B = 10$ and evaluate the performance on the CIFAR10 dataset. From our experiments, we observe that even with fewer clients at $C = 0.1$, and for other datasets, the performance of FHDnn is better than ResNet. Here, we present only the results for the settings mentioned earlier to keep it concise.

\textbf{Packet Loss}   As shown in Fig.~\ref{fig:loss}a, if the packet loss rate is extremely small, e.g., below $10^{-2}$, ResNet has very minimal accuracy loss. However, for more, realistic packet loss rates such as $20\%$ the CNN model fails to converge. When there is packet loss, the central server replaces the model weights from the lost packets with zero values. For example, $20\%$ packet loss rate implies $20\%$ of the weights are zero.  Moreover, this loss is accumulative as the models are averaged during each round of communication thereby giving the CNNs no chance of recovery. In contrast, FHDnn is highly robust to packet loss with almost no loss in accuracy. For FHDnn, since the data is distributed uniformly across the entire hypervector, a small amount of missing data is tolerable. However, since CNNs have a more structured representation of data with interconnections between neurons, the loss of weights affects the performance of subsequent layers which is detrimental to its performance.

\textbf{Gaussian Noise}  We experiment with different Signal-to-Noise Ratios (SNR) to simulate noisy links, illustrated in Fig.~\ref{fig:loss}b. Even for higher SNRs such 25dB the accuracy of ResNet drops by $~8\%$ under Non-IID data distribution. However it's more likely that IoT networks operating on low-power wireless networks will incur lower SNRs. For such scenarios, FHDnn outperforms ResNet as the latter fails to perform better than random classification. ResNet performance starts to completely deteriorate around 10dB SNR. The accuracy of FHDnn only reduces by $~3\%$, even at -10dB SNR, which is negligible compared to ResNet. 

\textbf{Bit Errors}  Fig.~\ref{fig:loss}c shows that CNNs completely fail when  bit errors are present. ResNet achieves the equivalent of random classification accuracy even for small bit errors. Since the weights of CNNs are floating point numbers, a single bit flip can significantly change the value of the weights. This, compounded with federated averaging, hinders convergence. We observe FHDnn incurs an accuracy loss as well, achieving $~72\%$ for IID and $~69\%$ for Non-IID data. FHDnn uses integer representations which is again susceptible to large changes from bit errors to some extent. However, our quantizer method with scaling described in Section~\ref{sec:unreliablecomm} assuages the remaining error.

\subsection{FHDnn Communication Efficiency}
So far we have benchmarked the accuracy of FHDnn for various network conditions. In the following, we demonstrate the communication efficiency of FHDnn compared to ResNet. We compare the amount of data transmitted for federated learning to reach a target accuracy of $80\%$. The amount of data transmitted by one client is calculated using the formula $data_{transmitted} = n_{rounds} \times update_{size}$, where $n_{rounds}$ is the number of rounds required for convergence by each model. The update size for ResNet with 11M parameters is 22MB while that of FHDnn is 1MB making it 22$\times$ smaller. From Section \ref{sec:rel_acc} we know that FHDnn converges 3$\times$ faster than ResNet bringing its total communication cost to 25MB. ResNet on the other hand uses up 1.65GB of data to reach the target accuracy.

In Fig.~\ref{fig:convergence}, we illustrated that FHDnn can converge to the optimal accuracy in much fewer communication rounds. However, this improvement is even higher in terms of the actual \textit{clock time} of training. We assume that federated learning takes places over LTE networks where SNR is 5dB for the wireless channel. Each client occupies 1 LTE frame of 5MHz bandwidth and duration 10ms in a time division duplexing manner. For error-free communication, the traditional FL system using ResNet can support up to 1.6 Mbits/sec data rate, whereas we admit errors and communicate at a rate of 5.0 Mbits/sec. Under this setting and for the same experiment as in Section 4.2, FHDnn converges in 1.1 hours for CIFAR IID and 3.3 hours for CIFAR Non-IID on average. On the other hand, ResNet converges in 374.3 hours for both CIFAR IID and CIFAR Non-IID on average.

\subsection{FHDnn: Effect of Communication Efficiency Strategies}
Even though FHDnn is much smaller than CNN models and the training converges faster, it's communication efficiency can be further improved. We use the MNIST dataset and the parameters from Section~\ref{sec:rel_acc} for our experiments.  Table~\ref{tab:results2} shows the final accuracy after 100 rounds of training and the improvement in communication cost for the respective approaches.

\begin{table}[h]
\centering
\label{tab:results2}
\caption{Simulation results }
\begin{tabular}{ccc}
\toprule
 \textbf{Method} & \textbf{Final Accuracy} & \textbf{Improvement} \\
    \hline
Baseline & 94.1\% & -  \\
    \hline
Binarized Differential Transmission & 91.2\% & 32x \\
    \hline
50\% Subsampling & 91.1\% &  2x \\
    \hline
50\% Sparsification \& Compression & 90.0\% & 2x \\
    \hline
10\% Subsampling & 90.7\% &  10x \\
    \hline
90\% Sparsification \& Compression & 91.6\% & 10x \\
   \bottomrule
\end{tabular}
\end{table}

The differential transmission approach binarizes the model difference, going from 32-bit floating point to 1-bit binary transmission to reduce the communication cost by 32x. For subsampling and sparsification \& compression approaches, the communication improvement depends on the percentage of the model values that are subsampled or sparsified. For example, if we subsample 10\% of the model, than the communication cost is reduced by 10x. Or, if we sparsify the model by 90\%, the reduction is 10x again. We present the final accuracy and improvement in communication cost at different subsampling and sparsification percentages.

\section{Conclusion}
In this paper we introduced methods to implement federated learning using hyperdimensional computing to enable communication efficient and robust federated learning for IoT networks. We first formalize the theoretical aspects of hyperdimensional computing to perform federated learning, presented as our first contribution called FedHD. To combat the inability of HDC to extract relevant features which consequently leads to poor performance of FedHD on large image classification, we propose FHDnn. FHDnn complements FedHD with a fixed contrastive learning feature extractor to compute meaningful representations of data that helps the HDC model better classify images. We described the federated hyperdimensional computing architecture, described the training methodology and evaluated FedHDC and FHDnn through numerous experiments in both reliable and unreliable communication settings. The experiment results indicate that FHDnn converges 3$\times$ faster, reduces communication costs by 66$\times$, local client compute and energy consumption by ~1.5 - 6$\times$ compared to CNNs. It is robust to bit errors, noise, and packet loss. Finally, we also showed that the communication efficiency of FedHDC and FHDnn can be further improved up to 32$\times$ with a minimal loss in accuracy.

\section*{Acknowledgements}
This work was supported in part by CRISP, PRISM and CoCoSys, centers in JUMP1.0 and 2.0 (SRC programs sponsored by DARPA), SRC Global Research Collaboration grants (GRC TASK 3021.001), and NSF grants \#2003279, \#1911095, \#1826967, \#2100237, and \#2112167.


%

\appendices
\section{Proof of Theorem 1}
\subsection{Additional Notation}
In our original analysis in Section~\ref{sec:fedhdc}, we used the variable $t$ to denote the communication rounds. Here, with a slight abuse of notation, we change the granularity of the time steps to be with respect to the SGD iterations instead of the communication rounds. Let $\mathbf{w}_{t}^{k}$ be the model maintained on the $k$-th device at the $t$-th step, and $\mathbf{w}_{t}$ be the global model. The clients communicate after $E$ local epochs for global aggregation. Let $\mathcal{I}_{E}$ be the set of those aggregation steps, i.e., $\mathcal{I}_{E} = \{nE \;| \; n=1,2,...\}$, so the client models are aggregated if $t+1 \in \mathcal{I}_{E}$. We introduce an additional variable $\mathbf{v}_{t+1}^{k}$ to represent the result of the SGD steps where no communication occurs, similar to~\cite{li2019convergence,stich2018local}. The following equation describes the local updates of the clients:

\begin{equation}
    \mathbf{v}_{t+1}^{k} = \mathbf{w}_{t}^{k}  - \eta_{t} \nabla F_{k}(\mathbf{w}_{t}^{k}, \xi_{t}^{k}) \nonumber 
\end{equation}

If $t+1 \notin \mathcal{I}_{E}$, we have $\mathbf{w}_{t+1}^{k} = \mathbf{v}_{t+1}^{k}$ because there is no communication and aggregation of models. On the other hand, if $t+1 \in \mathcal{I}_{E}$, the randomly selected clients $k \in \mathcal{S}_{t+1}$ communicate their models which are then aggregated and averaged at the server. This is summarized by the equation below.
\begin{align}
    \label{eq:classifier3}
    \mathbf{w}_{t+1}^{k} = \left\{
                \begin{array}{ll}
               \mathbf{v}_{t+1}^{k}, \;\;\;\;\;\;\;\;\;\;\;\;\;\;\;\;\;\;\;\;\;\;\;\;\,\,\,\, \text{if } t+1 \notin \mathcal{I}_{E}\\
                \frac{1}{|\mathcal{S}_{t+1}|}\sum_{k \in \mathcal{S}_{t+1}}\mathbf{v}_{t+1}^{k}, \;\;\; \text{if } t+1 \in \mathcal{I}_{E}
                \end{array}
              \right. \nonumber 
\end{align}

The variable $\mathbf{w}_{t+1}^{k}$ can be interpreted as the model obtained directly after the communication steps. We define two virtual sequences $\overline{\mathbf{v}}_{t+1} = \sum_{k=1}^{N}p_{k}\mathbf{v}_{t}^{k}$  and $\overline{\mathbf{w}}_{t+1} = \sum_{k=1}^{N}p_{k}\mathbf{w}_{t}^{k}$. Notice that if we apply a single step of SGD to $\overline{\mathbf{w}}$, we get $\overline{\mathbf{v}}_{t+1}$. These sequences are denoted virtual because both are not available when $t+1 \notin \mathcal{I}_{E}$ and we can only access $\overline{\mathbf{w}}_{t+1}$ when $t+1 \in \mathcal{I}_{E}$. We also define $\overline{\mathbf{g}}_{t} = \sum_{k=1}^{N}p_{k} \nabla F_{k}(\mathbf{w}_{t}^{k})$ and $\mathbf{g}_{t} = \sum_{k=1}^{N}p_{k} \nabla F_{k}(\mathbf{w}_{t}^{k}, \xi_{t}^{k})$ for convenience of notation. Therefore, $\overline{\mathbf{v}}_{t+1} = \overline{\mathbf{w}}_{t} - \eta_{t}\mathbf{g}_{t}$ and $\mathbb{E}\|\mathbf{g}_{t}\| = \overline{\mathbf{g}}_{t}$.

There are two sources of randomness in the following analysis. One results from the stochastic gradients and the other is from the random sampling of devices. To distinguish them, we use the notation $\mathbb{E}_{\mathcal{S}_{t}}(\cdot)$ when we take expectation over the randomness of stochastic gradients.

\subsection{Lemmas}
We present the necessary lemmas that we use in the proof of Theorem 1. These lemmas are derived and established in~\cite{li2019convergence}, so we defer their proofs.

\textit{Lemma 1} (Result of One Step SGD). \textit{Assume Properties 1 and 2 hold. If $\eta_{t} \leq \frac{1}{4L}$, we have}

\begin{align}
    &\mathbb{E}\|\overline{\mathbf{v}}_{t+1} - \mathbf{w}^{*}\|^{2} \leq (1- \eta_{t}\mu)\mathbb{E}\|  \overline{\mathbf{w}}_{t} - \mathbf{w}^{*} \|^{2} \nonumber \\ +  & \eta_{t}^{2}\mathbb{E}\|\mathbf{g}_{t} - \overline{\mathbf{g}}_{t}\|^{2} + 6L \eta^{2}_{t}\Gamma + 2\mathbb{E}   \Big[\sum_{k=1}^{N}p_{k} \|\overline{\mathbf{w}}_{t} - \mathbf{w}^{k}_{t} \|^{2} \Big]. \nonumber 
\end{align}

\textit{Lemma 2} (Bounding the Variance). \textit{Assume Property 3 holds. It follows that}
\begin{equation}
    \mathbb{E}\|\mathbf{g}_{t} - \overline{\mathbf{g}}_{t}\|^{2} \leq \sum_{k=1}^{N} p_{k}^{2}\sigma_{k}^{2}. \nonumber
\end{equation}

\textit{Lemma 3} (Bounding the Divergence of $\mathbf{w}_{t}^{k}$). \textit{Assume Property 4 holds, $\eta_{t}$ is non-increasing and $\eta_{t} \leq 2\eta_{t+E}$ for all $t\leq 0$. It follows that}
\begin{equation}
    \mathbb{E}\Big[ \frac{1}{N} \sum_{k=1}^{N} \| \overline{\mathbf{w}}_{t} - \mathbf{w}^{k}_{t}   \|^{2} \Big] \leq 4 \eta_{t}^{2}(E-1)^{2}G^{2}. \nonumber
\end{equation}

\textit{Lemma 4} (Unbiased Sampling). \textit{If $t+1 \in \mathcal{I}_{E}$, we have}
\begin{equation}
    \mathbb{E}_{\mathcal{S}_{t}} [\overline{\mathbf{w}}_{t+1}] = \overline{\mathbf{v}}_{t+1}  \nonumber
\end{equation}

\textit{Lemma 5} (Bounding the Variance of $\mathbf{w}_{t}^{k}$). \textit{For $t+1 \in \mathcal{I}_{E}$, assume that $\eta_{t}\leq 2\eta_{t+E}$ for all $t\geq0$. We then have} 
\begin{equation}
    \mathbb{E}_{\mathcal{S}_{t}}\|\overline{\mathbf{v}}_{t+1} - \overline{\mathbf{w}}_{t+1}\|^{2} \leq  \frac{N-K}{N-1}\frac{4}{K}\eta_{t}^{2}E^{2}G^{2}   \nonumber
\end{equation}

\subsection{Theorem 1}
We have $\overline{\mathbf{w}}_{t+1} = \overline{\mathbf{v}}_{t+1}$ whether $t+1 \in \mathcal{I}_{E}$ or $t+1 \notin \mathcal{I}_{E}$. Then, we take the expectation over the randomness of stochastic gradient and use Lemma 1, Lemma 2, and Lemma 3 to get

\begin{align}
    &\mathbb{E}\|\overline{\mathbf{w}}_{t+1} - \mathbf{w}^{*}\|^{2} = \mathbb{E}\|\overline{\mathbf{v}}_{t+1} - \mathbf{w}^{*}\|^{2}  \nonumber \\
    & \leq (1- \eta_{t}\mu)\mathbb{E}\|  \overline{\mathbf{w}}_{t} - \mathbf{w}^{*} \|^{2} + \eta_{t}^{2}\mathbb{E}\|\mathbf{g}_{t} - \overline{\mathbf{g}}_{t}\|^{2} \nonumber \\
    & \;\;\;\;+  6L \eta^{2}_{t}\Gamma + 2\mathbb{E}   \Big[\sum_{k=1}^{N}p_{k} \|\overline{\mathbf{w}}_{t} - \mathbf{w}^{k}_{t} \|^{2} \Big] \nonumber \\
    & \leq (1- \eta_{t}\mu)\mathbb{E}\|  \overline{\mathbf{w}}_{t} - \mathbf{w}^{*} \|^{2} \nonumber \\
    & \;\;\;\; + \eta_{t}^{2}\Big[\sum_{k=1}^{N} p_{k}^{2}\sigma_{k}^{2} + 6L\Gamma + 8(E-1)^{2}G^{2}\Big].\nonumber 
\end{align}

If $t+1 \in \mathcal{I}_{E}$, note that
\begin{align}
    \|\overline{\mathbf{w}}_{t+1} - \mathbf{w}^{*}\|^{2} 
    & =     \|\overline{\mathbf{w}}_{t+1} - \overline{\mathbf{v}}_{t+1} + \overline{\mathbf{v}}_{t+1} - \mathbf{w}^{*}\|^{2} \nonumber \\
    & = \underbrace{\|\overline{\mathbf{w}}_{t+1} - \overline{\mathbf{v}}_{t+1}\|^{2}}_{A_1} + \underbrace{\|\overline{\mathbf{v}}_{t+1} - \mathbf{w}^{*}\|^{2}}_{A_2} \nonumber \\
     & + \underbrace{\langle \overline{\mathbf{w}}_{t+1} - \overline{\mathbf{v}}_{t+1}, \overline{\mathbf{v}}_{t+1} - \mathbf{w}^{*}\rangle}_{A_3} . \nonumber 
\end{align}
When the expectation is taken over $\mathcal{S}_{t+1}$, the term $A_{3}$ vanishes because $\mathbb{E}_{\mathcal{S}_{t+1}}[\overline{\mathbf{w}}_{t+1} - \overline{\mathbf{v}}_{t+1}]=0$, that is, $\overline{\mathbf{w}}_{t+1}$ is unbiased. If $t+1 \notin \mathcal{I}_{E}$, $A_{1}$ is vanished as $\overline{\mathbf{w}}_{t+1} = \overline{\mathbf{v}}_{t+1}$. $A_{2}$ can be bounded using Lemmas 1 to 3 and Lemma 5. It follows that
\begin{align}
& \mathbb{E}\|\overline{\mathbf{w}}_{t+1} - \mathbf{w}^{*}\|^{2} \leq (1- \eta_{t}\mu)\mathbb{E}\|  \overline{\mathbf{w}}_{t} - \mathbf{w}^{*} \|^{2} \nonumber \\
& + \eta_{t}^{2}\Big[\sum_{k=1}^{N} p_{k}^{2}\sigma_{k}^{2} + 6L\Gamma + 8(E-1)^{2}G^{2} \Big].  \nonumber
\end{align}

If $t+1 \in \mathcal{I}_{E}$, the term $A_{1}$ can be additionally bounded using Lemma 5, then

\begin{align}
& \mathbb{E}\|\overline{\mathbf{w}}_{t+1} - \overline{\mathbf{v}}_{t+1}\|^{2} + \mathbb{E}\|\overline{\mathbf{v}}_{t+1} - \mathbf{w}^{*}\|^{2} \nonumber \\
& \leq (1- \eta_{t}\mu)\mathbb{E}\|  \overline{\mathbf{w}}_{t} - \mathbf{w}^{*} \|^{2} \nonumber \\
& + \eta_{t}^{2}\Big[\sum_{k=1}^{N} p_{k}^{2}\sigma_{k}^{2} + 6L\Gamma + 8(E-1)^{2}G^{2} + \frac{N-K}{N-1}\frac{4}{K}E^{2}G^{2} \Big].  \nonumber
\end{align}

Now, let $\Delta_{t} = \|\overline{\mathbf{w}}_{t+1} - \mathbf{w}^{*}\|^{2}$ for notational convenience. Also, let
\begin{equation}
    B = \sum_{k=1}^{N} p_{k}^{2}\sigma_{k}^{2} + 6L\Gamma + 8(E-1)^{2}G^{2} + \frac{N-K}{N-1}\frac{4}{K}E^{2}G^{2}.\nonumber
\end{equation}
We use a diminishing learning rate with $\eta_{t} = \frac{\beta}{t+\gamma}$ for some $\beta \geq \frac{1}{\mu}$ and $\gamma > 0$ such that $\eta_{1} \leq \text{min} \{\frac{1}{\mu},\frac{1}{4L} \} = \frac{1}{4L}$ and $\eta_{t} \leq 2\eta_{t+E}$. Now, we prove by induction that
\begin{equation}
     \Delta_{t} \leq \frac{v}{\gamma + t} \nonumber
\end{equation}
where
\begin{equation}
   v = \text{max}\{\frac{\beta^{2}B}{\beta \mu -1}, (\gamma+1)\Delta_{0}    \}. \nonumber
\end{equation}
The definition of $v$ ensures that it holds for $t =0$. If we assume the result holds for some $t>0$, it follows
\begin{align}
     \Delta_{t+1}  & \leq  (1-\eta_{t}\mu)\Delta_{t} + \eta_{t}^{2}B \nonumber \\
    &  = \Big( 1 - \frac{\beta \mu}{t + \gamma}  \Big)\frac{v}{t+\gamma} + \frac{\beta^{2}B}{(t+\gamma)^{2}} \nonumber \\
    & = \frac{t+\gamma-1}{(t+\gamma)^{2}}v + \Big[\frac{\beta^{2}B}{(t+\gamma)^{2}} - \frac{\beta \mu -1}{(t+\gamma)^{2}}v \Big] \nonumber \\
    & \leq \frac{v}{t+\gamma +1}. \nonumber
\end{align}
Then, by the strong convexity of $F(\cdot)$,
\begin{equation}
    \mathbb{E}[F(\overline{\mathbf{w}}_{t})] - F^{*} \leq \frac{L}{2}\Delta_{t} \leq \frac{L}{2}\frac{v}{\gamma+t}.\nonumber
\end{equation}
If we choose the parameters as $\beta = \frac{2}{\mu}$, $\gamma = \text{max}\{8\frac{L}{\mu}-1,E\}$ and define $\kappa = \frac{L}{\mu}$, then $\eta_{t} = \frac{2}{\mu}\frac{1}{\gamma+t}$. Using the fact that $\text{max}\{a,b\}\leq a+b$, we have
\begin{align}
       v & \leq \frac{\beta^{2}B}{\beta \mu -1}+  (\gamma+1)\Delta_{0}  \nonumber \\
       &  = 4\frac{B}{\mu^{2}} + (\gamma+1)\Delta_{0} \nonumber \\
       & \leq 4\frac{B}{\mu^{2}} + \Big(8\frac{L}{\mu}-1+E+1 \Big)\Delta_{0} \nonumber \\
       & =  4\frac{B}{\mu^{2}} + \Big(8\frac{L}{\mu}+E\Big)\|\overline{\mathbf{w}}_{1} - \mathbf{w}^{*}\|^{2}. \nonumber
\end{align}
Therefore,
\begin{align}
    \mathbb{E}[F(\overline{\mathbf{w}}_{t})] {-} F^{*} &  \leq \frac{L}{2(\gamma+t)} \Big[4\frac{B}{\mu^{2}} + \Big(8\frac{L}{\mu}+E\Big)\|\overline{\mathbf{w}}_{1} - \mathbf{w}^{*}\|^{2} \Big]  \nonumber \\
    & = \frac{2\kappa}{\gamma+t}\Big[ \frac{B}{\mu} + \Big(2L + \frac{E\mu}{4}\Big) \|\overline{\mathbf{w}}_{1} - \mathbf{w}^{*}\|^{2}\Big].
\end{align}

\ifCLASSOPTIONcaptionsoff
  \newpage
\fi

\bibliographystyle{ieeetr}
\bibliography{main}

\end{document}